# Towards Social Autonomous Vehicles: Efficient Collision Avoidance Scheme Using Richardson's Arms Race Model

Faisal Riaz[1], Muaz Niazi[2]
[1]Dept. Of Computing-Iqra University, Islamabad, Pakistan
[2]Dept. Of Computer Sciences-COMSATS, Islamabad, Pakistan
Email: [1]fazi_ajku@yahoo.com, *muaz.niazi@gmail.com

## Abstract

### Background

Road collisions and casualties pose a serious threat to commuters around the globe. Autonomous Vehicles (AVs) aim to make the use of technology to reduce the road accidents. However, the most of research work in the context of collision avoidance has been performed to address, separately, the rear end, front end and lateral collisions in less congested and with high inter-vehicular distances. Whereas the flock like topology, a typical scenario of urban traffic single one-way lane, where the traffic pattern is congested, the inter-vehicular distance is small and the chances of collisions is very high, has not been addressed sufficiently. The collision avoidance capabilities of AVs have been improved by using different methodologies, however, human-inspired designs have not been explored in this context, especially the human brain parts that involve in human-human interaction that make them social and help them in understanding and adapting the behaviour of other humans.

### Purpose

The goal of this paper is to introduce the concept of a social agent, which interact with other AVs in social manners like humans are social having the capability of predicting intentions, i.e. mentalizing and copying the actions of each other, i.e. mirroring. The proposed social agent is based on a human-brain inspired mentalizing and mirroring capabilities and has been modelled for collision detection and avoidance under congested urban road traffic.

**Method**

A detailed literature review has been performed to find out the existing research that uses human social life techniques to avoid the collisions between vehicles. During the literature review, Richardson's arms race model has been found very near to our problem that can be applied in this paper after modifications to the AVs to have efficient collision avoidance capabilities in a flock like topology. Then, we designed our social agent having the capabilities of mentalizing and mirroring and for this purpose we utilized Exploratory Agent Based Modeling (EABM) level of Cognitive Agent Based Computing (CABC) framework proposed by Niazi and Hussain (1). In the next step, we modified the differential equations defined by Richardson's arms race model to emulate the concept of mentalizing and mirroring in the proposed social agent. Then, the simulation model has been designed that helps in testing the behaviour of AVs in terms of collision avoidance using the proposed social agent based model and random walk. Then, using the behaviour space tool of NetLogo simulator, the performance of the proposed social agent based scheme has been computed and its results have been compared with the random walk based collision avoidance scheme. Furthermore, the practical validation of the proposed agent has been performed by building the proposed social agent inspired prototype AV. For the practical validation, a real road flock like topology has been created with the help of human driven motorcycles. In the last, the performance of proposed research work has been compared with the existing state-of-the-art.

**Results**

Our simulation and practical experiments reveal that by embedding Richardson's arms race model within AVs, collisions can be avoided while travelling on congested urban roads in a flock like topologies. The performance of the proposed social agent has been compared at two different levels. First at simulation level, the performance of the proposed social agent is compared using extensive experiment sets with Random walk based collision avoidance strategy and it has been found that proposed social agent based collision avoidance strategy is 78.52 % efficient than Random walk based collision avoidance strategy in a congested flock like topologies. Then the practical validation is performed in terms of efficiency and the results confirmed that the proposed scheme can avoid rear end and lateral collisions with the efficiency of 99.876 % as compared to the IEEE 802.11n based existing state of the art (33) mirroring neuron based collision avoidance scheme.

## 1. Introduction

Road collisions are an inevitable element of human life. According to (2), by 2030 road collisions will become the 5th major cause of human deaths. According to (3), road injuries are considered to be a twelfth main reason of human disability. According to (4), road collisions are the main cause of teen deaths and injuries. According to (5), road collisions are the second main reason for deaths in Europe. From these facts, it can be implied that road collisions cannot be avoided, but can be decreased using the latest advances in the field of Intelligent Transport System (ITS), like Autonomous Vehicles (AVs).

Autonomous vehicles can help in avoiding the road collisions. According to (6) AVs do not drink or distract like human drivers and have fewer chances of accidents as compared to the human-driven vehicles. In (2), researchers reported that AVs can lessen the collisions due to their better perception (e.g. no blind spot), decision making and faster execution of actuators like steering, brakes and gas pedals. Furthermore, (7), noted that the number of collisions can be decreased by introducing inter-AVs and Road Side Units (RSUs) based communication capabilities. (8) have mentioned that by using Google AVs there are fewer crashes as compared to the human-driven vehicles between the years of 2009 to 2015. However, the most of the research in the context of collision avoidance has been performed to address, separately, the rear end, front end and lateral collisions in less congested and with high intervehicular distances. Whereas the flock like topology, a typical scenario of urban traffic single one-way lane , where the traffic pattern is congested, the intervehicular distance is small, and the chances of the rear end, front end and lateral collisions is very high has not been addressed sufficiently. The collision avoidance capabilities of AVs have been improved by using different methodologies, however, human-inspired designs have not been explored in this context, especially the human brain parts that are involved in human-human interaction, which make them social and help them in understanding and adapting the behaviour of other humans.

Humans are social because of the specific brain structures. According to (9), humans use mentalizing and mirroring functions, imparted in their brains, to recognise and adapt the behaviours of other humans and hence make them social. The same has been reported by (10) that there are two neuron networks in the human brain which help humans to be social. The name of first neuron network is ventral medial prefrontal cortex (vmPFC) also known as mentalizing part, and the second network of neurons is known as Mirror Neuron System (MNS). The purpose of mentalizing part is to recognise the intention of other humans (10), whereas the MNS is responsible for helping a person to copy the actions of another person (11). It would be interesting to evaluate the mentalizing and mirroring concepts after incorporating them in the AVs to enhance their collision detection and avoidance capabilities, in a flock like topology, like humans interact with each other, know the intentions of each other, and avoid the conflicts by adapting some suitable strategies.

Now the question arises that what is the benefit of making AVs social. According to (12), agents are social when they share the same space. In our case when the AVs will travel in a flock like topology by sharing the highly congested urban road then they can be perceived as social agents and hence need some mechanisms that help them to avoid the collisions using human inspired social mechanism. However, to authors' best knowledge, AVs have not been designed yet as social agents. Furthermore, according to Libero et al. (13) the ability to interpret agents' intent of their actions is a vital skill in a successful social interaction and can be explored to enhance the pre-crash sensing capabilities of AVs. Furthermore, the capability to interpret the intent of other agents might be helpful in making proactive strategies to avoid the potential threats by making quick decisions in short reaction time. However, this line of research has not been also explored in the case of AVs that help them to be social and understanding the dangerous intents of other AVs and furthermore to avoid collisions.

Problem statement: AVs have not been designed as social agents, which have the capability to understand the intention of neighbouring AVs in a flock like topology and make collision avoidance manoeuvres by adapting their behaviours for safer road operations.

*Contribution*: In this paper, following contributions have been made.

- The architecture of social agent has been proposed.

- The concept of mentalizing and mirroring functions of human brain has been explored by using Exploratory Agent Based Modeling (EABM) level of Cognitive Agent Based Computing (CABC) framework (1) to tailor the components of the proposed social agent.
- A mathematical model using Richardson's Arms Race model (14) has been proposed to emulate the concept of mentalizing and mirroring components within the proposed social agent.
- Computer algorithms of mentalizing and mirroring have been proposed.
- UML design of simulation has been proposed.
- Simulation study of proposed technique has been performed.
- Extensive testing of proposed technique has been performed in comparison with random walk based travelling strategy.
- A prototype of social agent based AV has been developed.
- The proposed social agent has been deployed on a prototype AV platform and its performance has been validated in a flock like topology.

The rest of the paper is organised as follows. Section 2 presents the motivation behind this research work. Section 3 discusses our proposed methodology. Section 4 is the background. Section 5 is a literature review. Section 6 presents the proposed social agent architecture along the Richardson's Arm race-based mathematical modeling of its social components. A simulation environment, its UML design and test cases have been presented in 7. Section 8 presents the results and discussion of simulation experiments. Practical validation of the proposed social agent has been performed in section 9. The comparison with existing state of art has been made in section 10 and section 11 concludes the proposed research and points towards the future directions.

## 2. Motivation Behind The Research Work

After a detailed literature, we found that most of the research on collision avoidance has been performed to address the three types of scenarios presented in figure 1a, 1b, and 1c. The scenario presented in figure 1a is presenting rear end collision avoidance using on-board sensors such as sonars, Light Detection and Ranging (LIDAR), and cameras, whereas the scenario presented in figure 1b, is depicting the collision avoidance in the context of platooning and Adaptive Cruise Control (ACC) using cooperative communication approach. To address the scenario presented in figure 1a, many rear end collision avoidance solutions based on on-board sensors or wireless communication have been proposed. Gracia et al. (15), proposed sliding mode control based rear end collision avoidance solution. Van et al. (16), proposed rear end collision avoidance between vehicles using linear quadratic optimal control technique. Milanes et al. (17), proposed rear end collision detection and avoidance system using fuzzy logic. Sato and Akamatsu (18), modelled the human driver characteristics like driving style, reaction time and cognitive state using fuzzy logic to propose the rear end collision avoidance scheme. Li et.al have proposed GPS-enabled the vehicle to vehicle communication based rear end collision avoidance

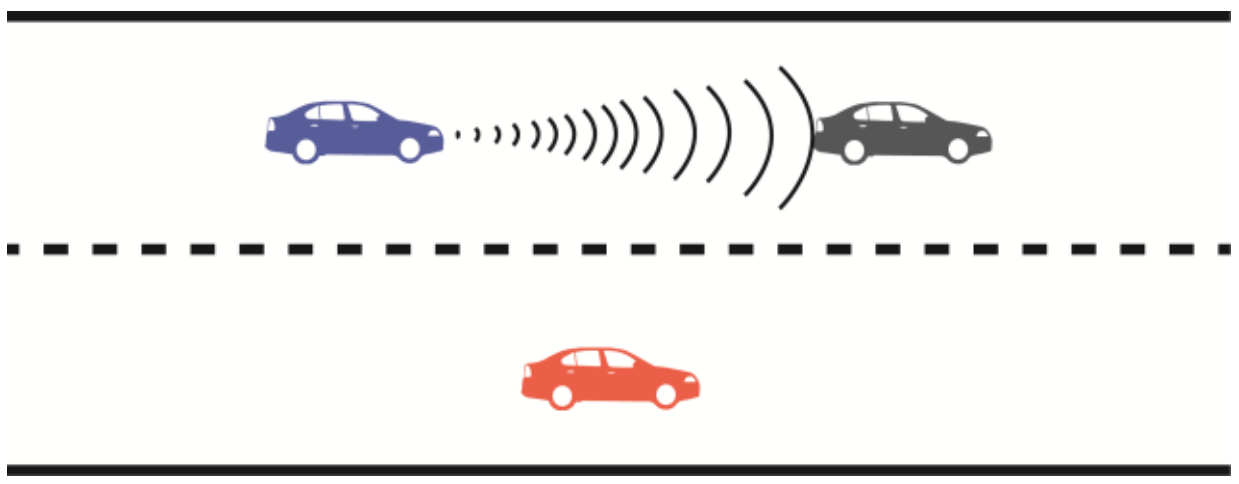

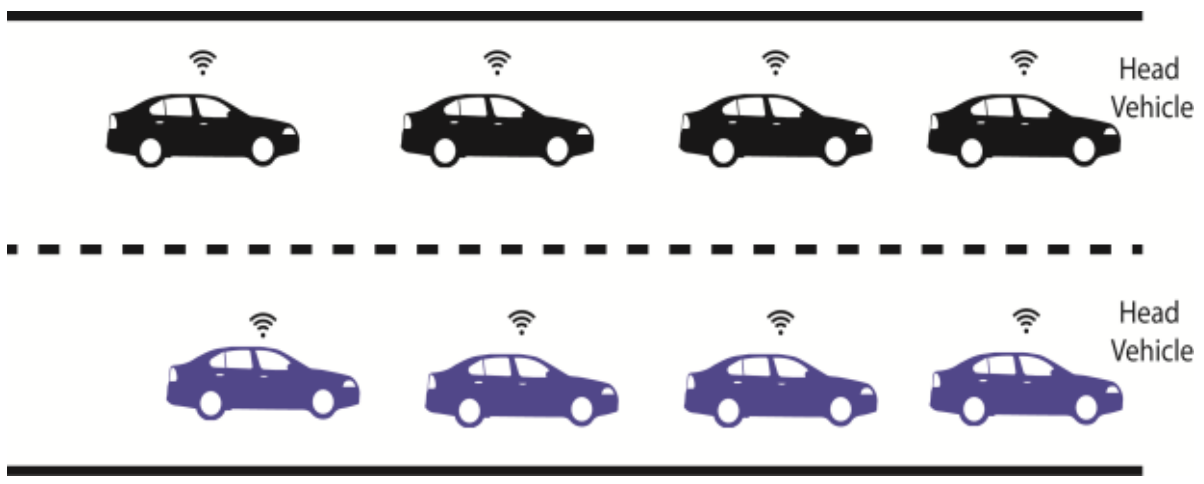

a b

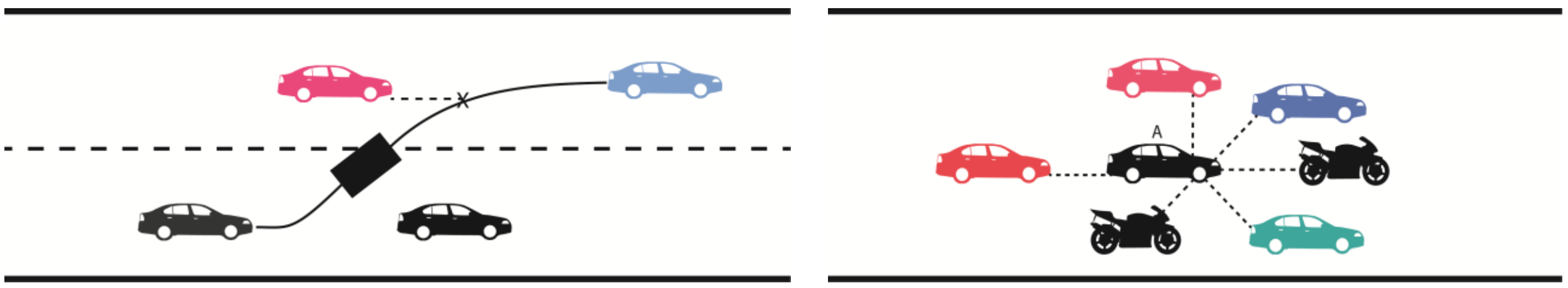

c d

Fig. 1 Collision avoidance scenarios addressed extensively in literature (a) Rear end collision avoidance, (b) Cooperative collision avoidance in Adaptive Cruise Control or Platooning, (c) Lane departure or Lateral collision avoidance, (d) Flock like topology (A typical scenario in congested urban road)

system in (19)[19](19)(Li, Lu et al. 2014)(Li, Lu et al. 2014)(Li, Lu et al. 2014)(Li et al. 2014). (Xiang et al. 2014), proposed GPS enabled rear end crash warning system using DSRC based inexpensive high-end devices. In the literature, researchers have addressed platooning and ACC, figure 1b, scenarios with extensive research work. In this regard, Liu and El Kamel (20) have proposed a decentralised cooperative adaptive cruise control algorithm using V2X communication. Milanés et al. (21) have proposed Cooperative Adaptive Cruise Control (CACC) in Real Traffic Situations using Vehicle-2-Vehicle (V2V) communication. In another research work, a vehicle platoon management frame work with the concept of the platoon leader and Vehicular Adhoc Network (VANET) has been proposed by Amoozadeh et al (22). The third main scenario, which has been addressed by various researchers, is lane departure/ lateral collision avoidance as shown in figure 1d. In (23)[23](23)(Schwindt, Kim et al. 2015)(Schwindt, Kim et al. 2015)(Schwindt, Kim et al. 2015)(Kim et al. 2015), a lane departure warning system is proposed using left, right, rear and forward sensors, a direction sensor, a processing unit, memory, and I\O interface. This system uses the front sensor to check the lane location of the vehicle and tracks the vehicle coming from the opposite side. The sensors of the vehicle help the vehicle in avoiding the collisions from the lateral vehicles during an overtaking manoeuvre. In another research work, the cognitive automatic overtaking system using vision system and fuzzy logic based controller is proposed to avoid the lateral collisions during overtaking manoeuvres by Milanes et al. (24) [24] (24) (Milanés, Llorca et al. 2012) (Milanés, Llorca et al. 2012) (Milanés, Llorca et al. 2012) (Milanés et al. 2012). Chu et al. (25) have proposed Dedicated Short Range Communication (DSRC) based lane departure and safe overtaking scheme for two-lane rural highways. However, the scenario depicted in the figure 1d has been ignored at large. The presented scenario depicts the vehicle travelling at low speed in congested urban traffic on single one-way road. If we see the details of the scenario, then it looks like a flock of vehicles, which are travelling in the same direction on a congested road and their ultimate goal is to reach their destinations safely. If we consider the vehicle A, figure 1d, as an autonomous vehicle, then it needs a robust motion controller, which helps AV to travel safely by avoiding front end, rear end and lateral collisions. The need of robustness is due to heavily congested traffic, which decreased the inter-vehicular distance to dangerous limits. So the awareness of neighbouring vehicles position and quick reaction time is the key to avoiding the collisions efficiently. During the literature review, we tried our best to find such a published work that address this issue with sufficient details and practical validation approach, but to our best knowledge, no such work has been reported. Then we analysed the above-mentioned research ideas, which have been done to address the scenarios 1a, 1b, and 1c but we found them unsuitable to address the scenario depicted in figure 1d. The rear end collision avoidance solutions provided to address the scenario 1a have following issues in this context. The mathematical based solutions provided by (15) and (16) are highly dependent on precise mathematical models as noted by (26) and have not been modeled by keeping in view the non-linear factors like road traffic pattern and driver reaction time. Whereas, the fuzzy logic based solutions, provided by (17) and (18) rely on the number of fuzzy rules and an excessive number of such will straightforwardly prejudice its efficiency in terms of delayed reaction time. The solutions provided for scenario 1b are using DSRC based V2V or V2I communication, whereas DSRC has been proved to be failed due to long packet delay and communication failure in congested urban traffic. In the same way, the solutions provided for scenario 1c are using fuzzy logic or wireless communication that are not suitable to address the problems associated with scenario 1d. Furthermore, all of the above-mentioned solutions address rear end, front end and lateral collisions separately. No such framework is available that helps the AV to avoid the rear end as well as lateral collisions at the same time with the quick reaction in the scenario of heavily congested urban traffic.

This lack of research gap motivates us to propose a novel scheme that helps the AV to travel in a flock like topology, which is very common in 3rd world countries such as Pakistan, India, Bangladesh and Srilanka.

## 3. Background

In this section, the short details of CABC framework have been provided.

### Cognitive Agent-Based Computing (CABC)

Agent-based modeling (ABM) and complex networks (CN) are two popular modeling tools for understanding Complex Adaptive System (CAS). In 2011, a unified framework named Cognitive Agent-based Computing (CABC) combining these two modeling paradigms was proposed by Muaz et al. (1) for the better understanding of CAS.  The CABC helps the cross-disciplinary researchers to develop the understanding of their area related CAS by using the different types of models. It provides guidelines to the multidisciplinary researchers regarding how they can develop computational models of CAS even they belong to social science, life science or computer science. The unified framework provides four understanding and development levels of CAS along with related case studies. The Complex Network Modeling, Exploratory Agent Based Modeling (EABM), Descriptive Agent Based Modeling (DREAM), and Validated Agent-Based modelling are the first, second, third and fourth levels defined under the CABC framwok respectively. However, in this paper, we have utilized only EABM to model our social agent. A short description about EABM is given as under.

**Exploratory Agent Based Modeling.** The second level of framework is Exploratory Agent Based Modeling (EABM). When the researchers are interested in extending existing ideas related to the agent based modeling belonging to the other fields, the EABM is a useful guideline paradigm in this regard. Using EABM, researchers can build experimental or proof of concepts, which help in defining the further scope and feasibility of the future research.

Under the guidelines of EABM, we have explored those human brain functions that help them to be social. Furthermore, using EABM, we have built both simulation and practical proof of the concept of the proposed collision avoidance scheme..

## 4. Method

Figure 2 presents the method that has been proposed to enhance the capabilities of AVs by using some human social scheme. A detailed literature review has been performed to find out the existing work, which uses human social life techniques to avoid the collisions between AVs. To our best knowledge, no such literature has been found in authentic research databases.   However, we found literature that supports the usage of human social life schemes to model, the robots in terms of social acceptability. During the literature review, Richardson's arms race model had been found very near to our problem, which can be applied after modifications to the AVs to have efficient collision avoidance capabilities in a flock like topology. In the next step, we modified the Richardson's arms race model according to our requirement. Then, we designed our agent having the capabilities of mentalizing and mirroring and for this purpose we utilized EABM. The detailed description of agent design is elucidated in section 4. Then, the simulation model has been designed, which helps in testing the behaviour of AVs in terms of collision avoidance using Richardson's arms race model and random walk. It is important to mention here that to give a comparative study, we have designed the random walk based simulation as well. The simulation parameter for the simulation has been selected carefully. Then, using the behaviour space tool of NetLogo simulator, the performance of the proposed social agent based scheme has been computed and its results have been compared with the random walk based collision avoidance scheme. Furthermore, the practical validation of the proposed agent has been performed by building the proposed social agent inspired prototype AV. For the practical validation, a real road flock like topology has been created with the help of human driven motorcycles. In the last, the performance of proposed research work has been compared with the existing state-of-the-art.

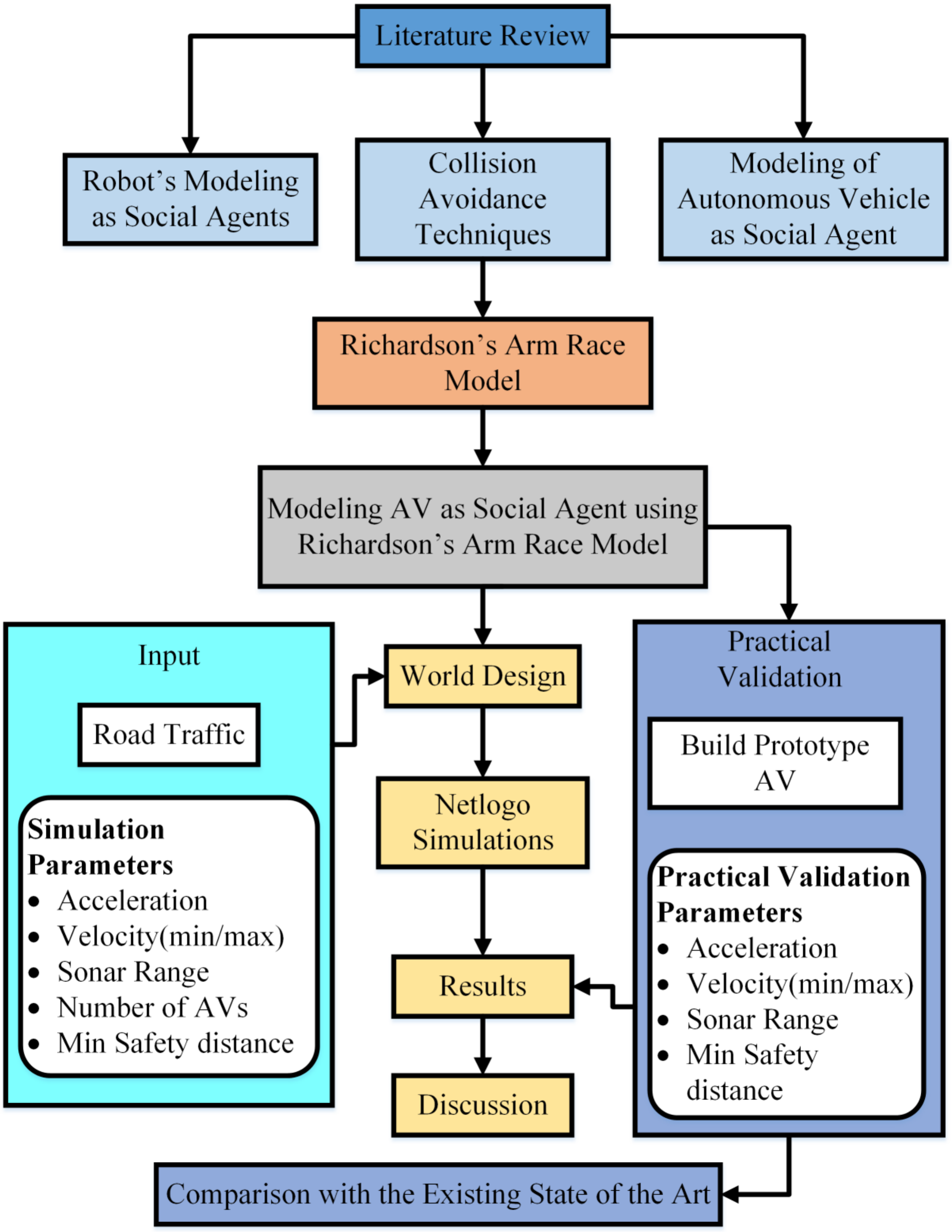

Fig. 2 Proposed Method

## 5. Literature Review

In the literature, artificial agents have been modelled as social entities for different applications. However, to the authors' best knowledge modelling and simulating the agents installed within the AV, as social entities for collision avoidance, is the first effort made in this paper.

Bicchi and Tamburrini (12) devised the collision avoidance mechanism in the artificial society of robots by making them social. Each robot keeps track of its neighbouring robots, same as humans follow social rules and avoid collisions in crowded spaces, and adapt collision avoidance strategy accordingly. Furthermore, the authors have suggested that such teams of robots can be build by following human social life protocols that help them to co-exist and move safely. According to (27) in near future, AVs will share the road with other road commuters and will become the part of a complex social-technical system. To be socially accepted in this complex socio-technical system, AVs need novel AV-X , X={ Human driven vehicles, pedestrians, other AVs }, interaction protocols. Furthermore, the authors have declared AVs as embodied intelligent agents. However, in this research work authors have just presented the theoretical concept of making AVs social and no practical steps have been taken in this regard. According to (28), a new generation of robots have a need for the social mechanisms that help them to engage the post-stroke patients in a better way. It has been noted by the authors that creating robots that has the capability to adapt their behaviour according to the personality of patients is a difficult task. In this regard, they have proposed a learning algorithm using policy gradient reinforcement learning (PGRL). The proposed algorithm first parameterized the behaviour of the post - stroke patient and then approximates the gradient of the reward function, and in the last help in taking steps towards a local optimum. Using these three steps, the experimental results have proven that the robots can change their behaviour according to the personality of the patients (28).

Kizilcec et al. (29) have evaluated the role of social robots as an online instructor, which teach the students through videos. To make the robots social, authors have equipped the robots with a voice mechanism along with facial expressions and body gestures. The experimental results prove that these social robots can be utilised as second best option to replace the human instructors. (30) and (31) employ the social forces model (32) where the agents are exposed to different repulsive and attractive forces depending on their relative distances.

# 6. Proposed Social Agent Architecture Using Exploratory Agent Based Modeling

As mentioned earlier in introduction section that our AV is designed inspired by the human capability of monitoring their neighbours and then adapting the same moves as their neighbours. We have utilized exploratory agent based modeling level of the CABC framework to explore the human brain inspired mechanism in the design of our social agent. The proposed agent is envisaged to avoid road accidents by keeping track of their neighbouring AVs and then performing the same manoeuvre as they do. The proposed agent possesses the ability to react in the event of danger inspired by human brain capable of mirroring and is proposed to be housed inside the vehicle. Recall that the agent is responsible for detecting potential threats and take necessary actions if required. The architecture of the proposed agent is presented in figure 3.

## 6.1 Description Of The Agent

It can be seen from the figure 3 that the proposed architecture consists of five main modules.

- **Sensory Module**: It keeps track of the distance between neighbouring cars on a road segment.
- **Mentalizing Module:** This module helps the AV to find the intention of neighbouring AVs. To find out the intentions, Richardson's arms race model equations 1 and 2, presented in section 6, have been employed. The mentalizing module keeps sensors data to find out the current motion pattern, which helps the AV to predict the potential collision threat in advance.

- **Mirroring Module:** The mirroring module helps the AV to change its trajectory according to the changed trajectory of the nearest AV. To create the capability of mirroring in AVs, we have utilized the equations 3 to 7 of the modified Richardson's arms race model.
- **Motor Module**: This module will initiate the execution module to execute the mirroring instructions, adapted angle and speed.
- **Execute Action Module**: This module will act in the place of the human driver to perform accident avoidance manoeuvre.

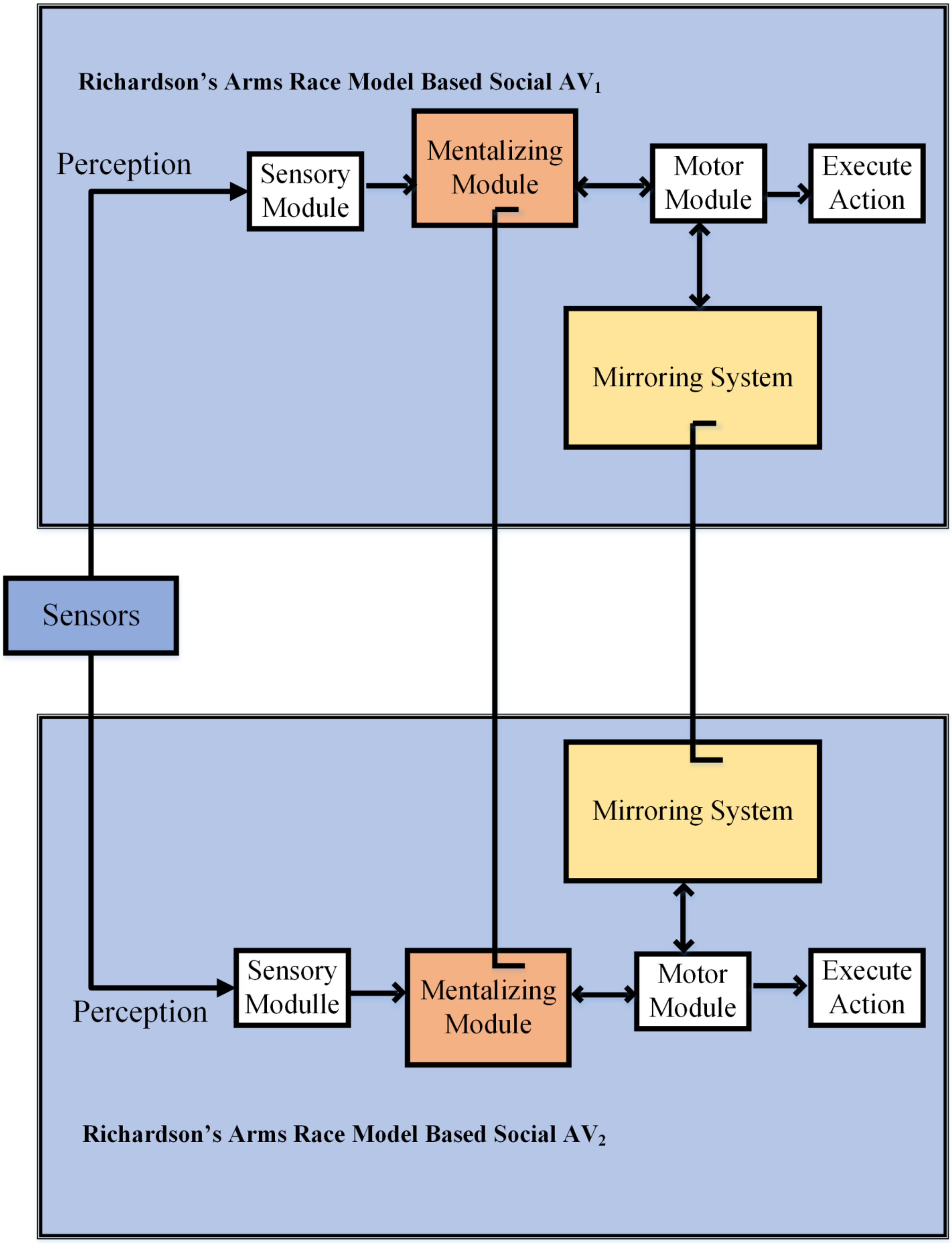

Fig. 3 AVs installed with proposed social agents interacting socially with each other

The figure 3, is presenting the interaction between two AVs using proposed agent architecture. Both AVs keep track of each other's movement intentions and avoid collisions using mirroring option.

## 6.2 Proposed Richardson's Arm Race-Based Mathematical Modelling For The Proposed Social Agent

It has been noted earlier in the previous sections that the proposed social agent incorporates the notion of intention understanding and adapting the behaviour of neighbouring AVs. In order to express, these capabilities of social agent mathematically the Richardson's arms race model is employed (14). The Richardson's model, which studies the circumstances under which two nations can avoid war, uses a set of linear differential equations. This work uses the said model to formulate the generation of fear in the proposed social agent. As shown in figure 3 both social agents of the two vehicles are exchanging their positions to evade the chance of an accident using distance-measuring sensors.

Consider two vehicles $v_1$ and $v_2$ that are moving on the road very close to each other. The position of $v_1$ at time n with respect to $v_2$ is represented by $v_1$ (n). Similarly, $v_2$ (n) represents the position of $v_2$ at the time n with respect to $v_1$. The change in the position of $v_1$ and $v_2$ with time shall be given by:

$$\Delta v_1(n) = v_1(n) - v_1(n-1) \quad (1)$$

$$\Delta v_2(n) = v_2(n) - v_2(n-1) \quad (2) \text{ (Mentalizing)}$$

Assume that while moving together on the same road, $v_2$ changes its relative position (for example while overtaking, etc.). In order to maintain a safe distance, $v_1$ shall have to change its position according to the change in $v_2$'s position. The equation 2 helps the proposed agent in performing its mantalizing function i.e. it helps in assessing the gesture, manoeuvre in this case, of the nearest neighbouring vehicle. After assessing the relative position of nearest neighbour, there will be a need to execute the safety manoeuvre. However, the question arises what should be the nature of safety manoeuvre. Here equation 3 comes, which helps the proposed social agent in performing its mirroring function. Hence,

$$\Delta v_1(n) = \delta_1 \Delta v_2(n\text{-}1) \quad (3) \text{ (Mirroring)}$$

Where $\delta_1$ is referred as the position coefficient.

Note that the change in the position of v1 is limited by the road width.

$$\Delta v_1(n) = \delta_1 v_2(n-1) - \alpha_1 v_1(n-1) \quad (4)$$

$$\Delta v_2(n) = \delta_2 v_1(n-1) - \alpha_2 v_2(n-1) \quad (5)$$

Where $\alpha_1$and $\alpha_2$ are positive constants, representing the road capacity limits in terms of performing safety manoeuvres. Note that the intensity of fear experienced by a vehicle also depends on its type and size and is represented by g in equations 6 & 7. A lighter vehicle will have the higher fear intensity and vice versa.

The goal of the vehicle, which is ultimately its safety, has been represented by h. Now the equations 4 & 5 can be written as:

$$\Delta v_1(n) = \delta_1 v_2(n-1) - \alpha_1 v_1(n-1) + g_1 * h_1 \quad (6) \text{ (Mirroring final equation)}$$

$$\Delta v_2(n) = \delta_2 v_1(n-1) - \alpha_2 v_2(n-1) + g_2 * h_2 \quad (7)$$

As we have seen, in Richardson's construction of the model the parameters δ1, α1, g and h have very special meanings, which suggested that these constants should be positive. However, it has since been argued that negative parameters can have equally relevant interpretations and that both mathematically and substantively it makes more sense to consider a general model in which parameters are not constrained [64]. We, therefore, rewrite (6) and (7) in a more standard form:

$$\Delta v_1(n) = \alpha_1 v_1(n-1) + \delta_1 v_2(n-1) + g_1 * h_1 \quad (8)$$

$$\Delta v_2(n) = \alpha_2 v_2(n-1) + \delta_2 v_1(n-1) + g_2 * h_2 \quad (9)$$

In addition, using equations 1 and 2, it can be written as

$$v_1(n) = (1+\alpha_1)v_1(n-1) + \delta_1 v_2(n-1) + g_1 * h_1 \quad (10)$$

$$v_2(n) = \delta_2 v_1(n-1) + (1+\alpha_2)v_2(n-1) + g_2 * h_2 \quad (11)$$

If we define

$$(1+\alpha_1) \equiv \beta_1, (1+\alpha_2) \equiv \beta_2$$

So

$$v_1(n) = \beta_1 v_1(n-1) + \delta_1 v_2(n-1) + g_1 * h_1 \quad (12)$$

$$v_2(n) = \delta_2 v_1(n-1) + \beta_2 v_2(n-1) + g_1 * h_1$$

We have shown the formulation of the model for two vehicles only. The model can be extended for N vehicles in the future.

## 6.3 Proposed Algorithm Design

In this section, two algorithms have been presented related to the agent-based implementation of Richardson's arms race model based collision avoidance system. The algorithm presented in figure 4 can be summarised as follows. The setup procedure helps in defining the two different types of AVs like the two nations of Richardson's arms race model. The colour of the first type of AVs is Red and the second type of AVs is black. The both types of vehicles set initially to the minimum velocity that can be changed using a slider, available on the simulation interface. The red AVS initial heading is set to the $90^o$ and the initial heading of black AVs is $120^o$. These different headings help in creating collision-leading situations. Furthermore, the operating environment has been defined as well, which is a ground consist of green patches.

The second algorithm is presented in figure 5. This algorithm helps in performing collision avoidance between AVs using Random walk pattern and Richardson's arms race model. This algorithm helps in implementing both the proposed approach and random walk travelling pattern. The algorithm can be summarised as follows. If the simulation mode is set to the ALLSOCIALAVs, then the AVs will use social agent model to avoid the collision, otherwise, the Random walk pattern will be employed by the AVs. In the case of social agent based model, vehicles will start their motion with current speed and consult the mentalizing procedure after a fix interval of time. The mentalizing procedure will create a list of all neighbours using the find-near mates procedure. Then among the near mates, the nearest neighbour will be marked and the safety distance will be computed. If the nearest neighbour creates a threat, then the mirroring procedure will be initiated and it will avoid the collision by adapting the angle and speed of the nearest neighbour.

In the case of Random walk pattern, the AVs will start their motion with the current speed and head 89, and then the AVs will change their heading to 200 degrees. Then the AVs attain maximum acceleration rate the first time and then switch to the low speed, and exhibit random behaviours.

## 7. Simulation

The details of the simulation has been discussed in this section. First of all, the proposed UML design of the simulation is presented in section 7.1. Then in section 7.2, the simulation environment has been discussed and in the last, the simulation parameters, along the test case design has been presented in section 7.2.1 and 7.2.2 respectively.

```
Data: N
Result:  set up the simulation Environment
Proc  setup
              set-default-shape turtles "car"
               create-turtles No_of_Red_AVs [ setup- Autonomous Vehicles ]
               create-turtles No_of_Black_AVs [ setup- Autonomous Vehicles
              ]
                ask patches
               [set pcolor green]
               watch turtle 0
                reset-ticks
end proc
Proc setup Autonomous Vehicles
               set color black, Red
               set xcor random-xcor
               set ycor random-ycor
               set heading  [90 and 120]
               set collision_done 0
               set no-of-collisions 0
               set random_behaviour  [true, false]
               set speed min_velocity
  end proc
```

Fig. 4 Algorithm 1: Setting up the simulation environment

## 7.1 Proposed Simulation UML Design

The UML design of the simulation has been presented using figure 6 to 9 that help in building a rigorous design of the proposed research. First, the use case diagram has been presented in figure 6. The use case diagram is depicting the main use cases that the AV will perform, in simulation, to execute the proposed approach.  "Compute danger using Richardson's arms race model" and "avoid danger" are the key use cases, which employee the mathematical models presented in section 5. It is important to mention, here, that the difference between "avoid danger" and "avoid collision " is that the first one is the logical thought and the former one is the practical implementation of first one by controlling actuators of the AV. In the next step, a sequence diagram of our proposed system has been modelled as presented in figure 7. The sequence diagram helps us in designing the verified logical sequence between the different components of the proposed system within the simulation.

```
Data: Richardsons' arms race model, local, and global variables
Result: Collision detection and avoidance
Main Proc ()
        ifelse scenarios ← "AllSocialAVs" then
          ask turtles
                  fd speed
                  danger
        else
          ask turtles
                  fd speed set heading random 89
                  fd speed set heading random 200
                  ifelse random_behaviour = true  then
                          set speed speed + max_acceleration
                          set random_behaviour false
                          if speed > max_velocity  then
                                  set speed max_velocity
                          end if
                  else
                          set speed speed + deceleration
                          set random_behaviour true
                          if speed < min_velocity then
                                  set speed min_velocity
                          end if
                  end if
                  count-collisions ;;; calling no of collisisons
               counting procedure for red AVs
          end if
        tick
          let total_no_of_collisions Total_no_of_collision
          let collision-per-time collision_per_time
          stop
end Proc
Mentalizing Proc ()
        to danger ()
          find-nearmates Proc ()
          if any? Nearmates then
                  find-nearest-neighbor
          end if
          if distance nearest-neighbor  <= minimum-safety_distance
        then
                  Mirroring Proc ()
                  Accelerate Proc ()
          end if
 end Proc
          find-nearmates Proc ()
                  set nearmates other turtles in-radius Sonar_Range
          end Proc
          find-nearest-neighbor Proc ()
                  set nearest-neighbor min-one-of nearmates [distance
                myself]
          end proc
Mirroring Proc ()

          set heading ([heading] of nearest-neighbor)
          decelerate
end proc
        decelerate Proc () ;; turtle procedure
                  set speed [speed] of nearest-neighbor - deceleration
```

Fig. 5 Algorithm 2: Richardson's arms race model inspired Agent-based collision detection and avoidance algorithm

The next diagram is the class diagram as presented in figure 8. The class diagram helps in defining the properties, variables in this case, and functions of the proposed system. In the last, activity diagram is presented in figure 9. From the figure 9, it can be seen that the AVs will initiate its sonar to have the surroundings information. From the input of the sonar, a list of neighbouring AVs will be generated and corresponding distances will be computed. If the distance of nearest neighbour will lie in the critical region, means equal to or less than the safety distance, then AV feels danger using the mathematical model of proposed Richardson's arms race model and in the result, AV will execute the collision avoidance manoeuvre. If the AV fails in avoiding the collision then the collision will be counted, otherwise, AV will keep processing the neighbouring AVs information using its sensors.

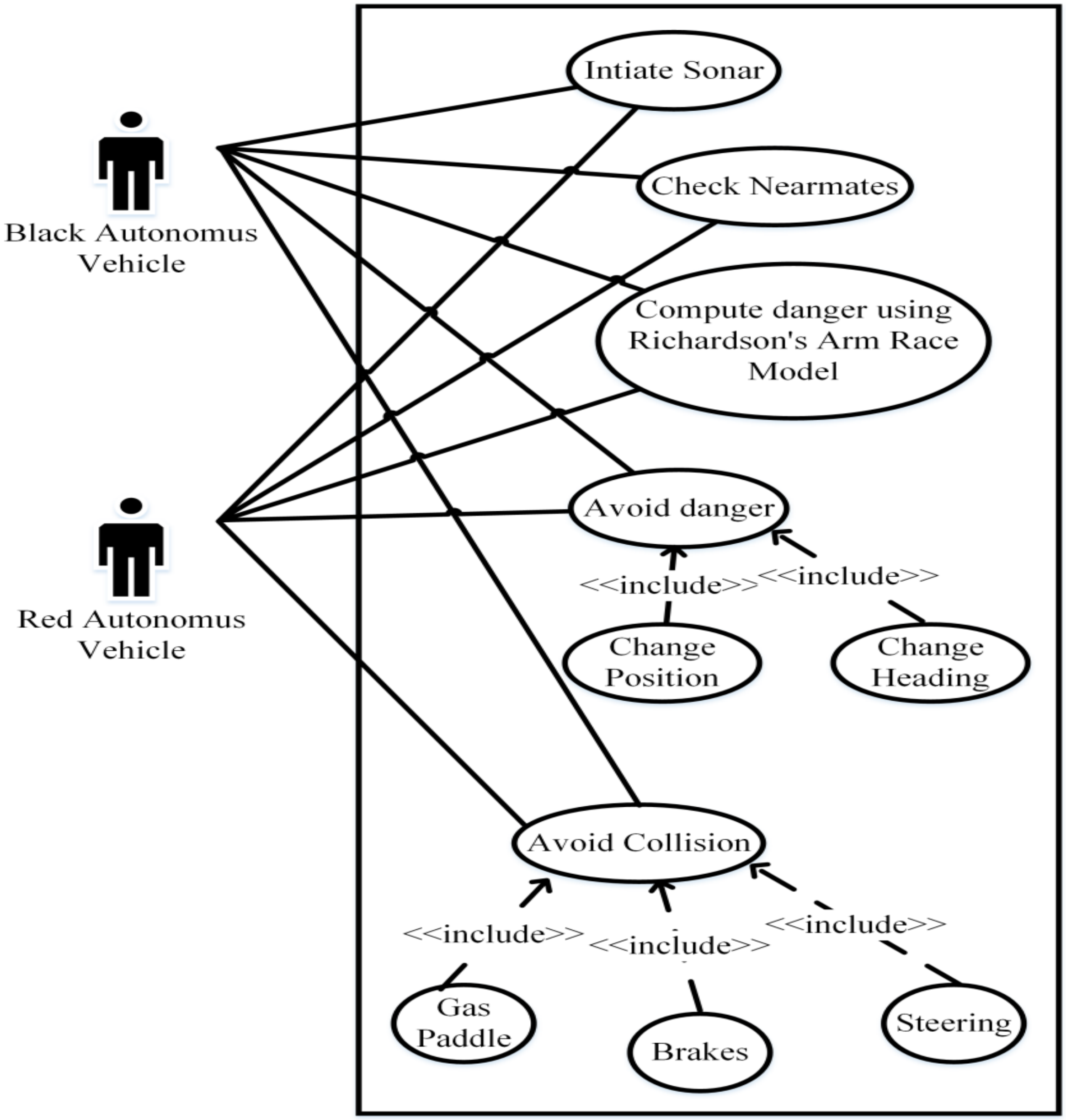

Fig. 6 Algorithm 2: Use Case diagram of the proposed Richardson's arms race model inspired Agent-based collision detection and avoidance scheme

**SEQUENCE DIAGRAM**

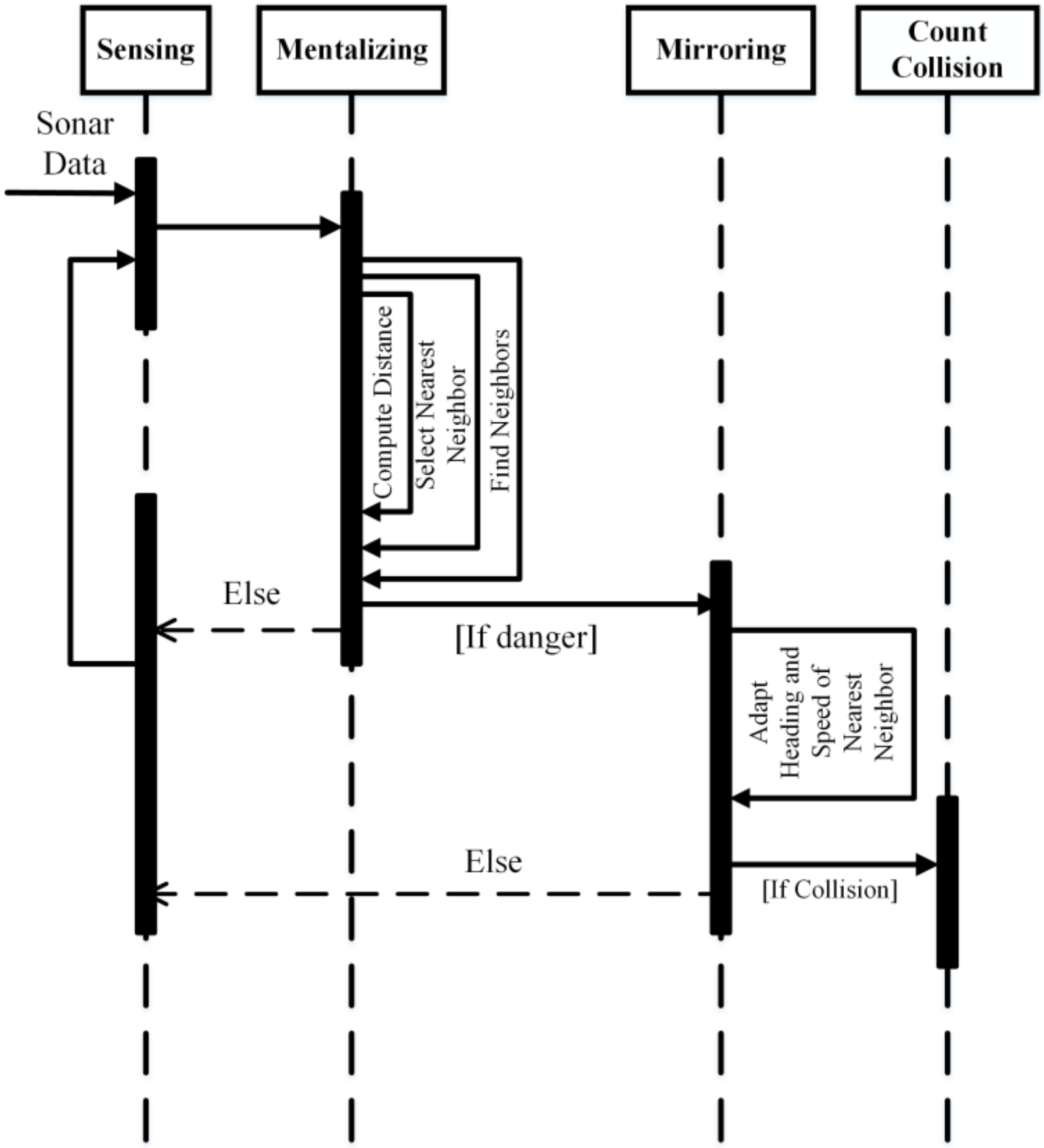

Fig. 7 Sequence diagram of the Richardson's arms race model inspired Agent-based collision detection and avoidance scheme

**CLASS DIAGRAM**

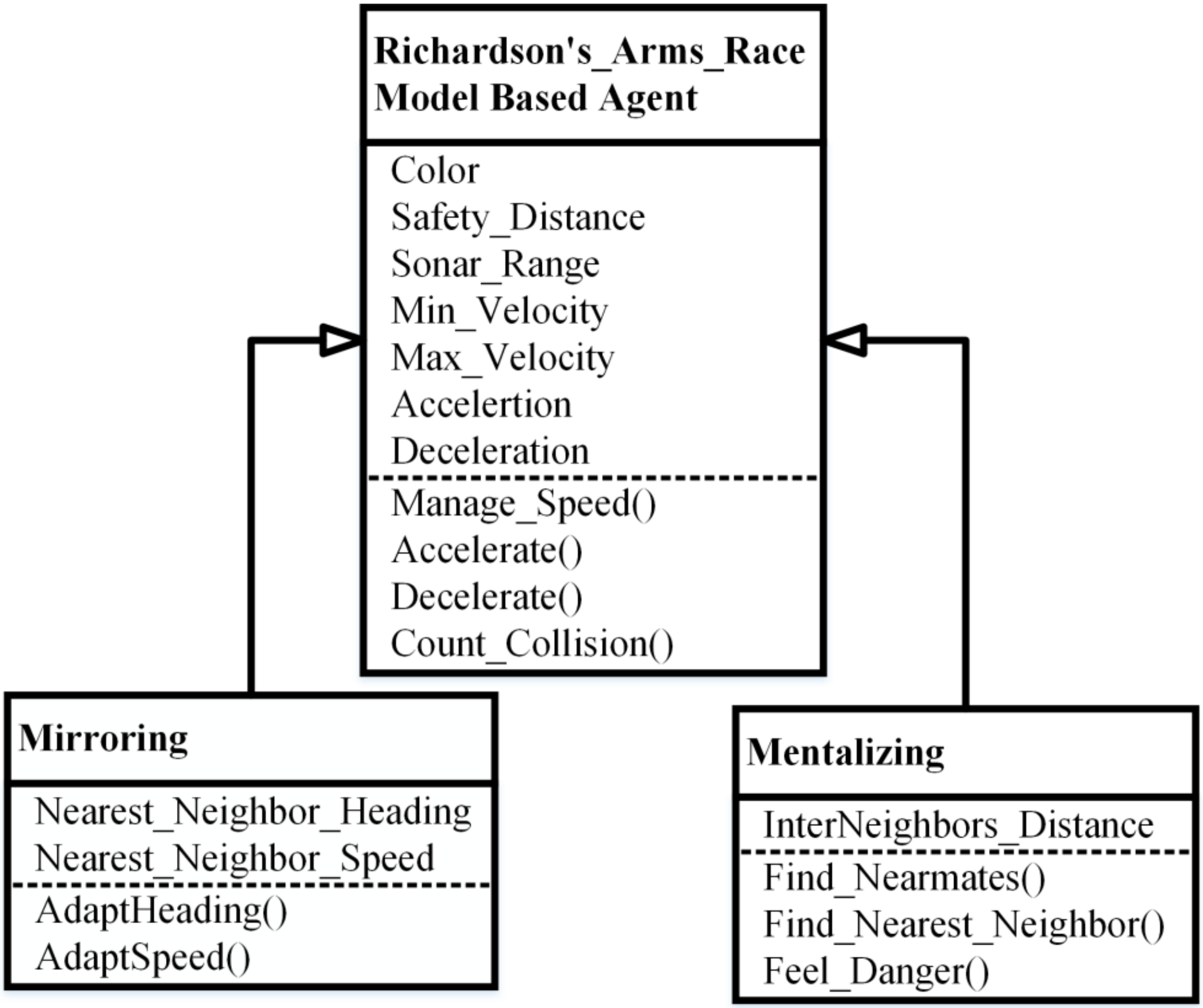

Fig. 8 Class diagram of the Richardson’s arms race model inspired Agent-based collision detection and avoidance scheme

**ACTIVITY DIAGRAM**

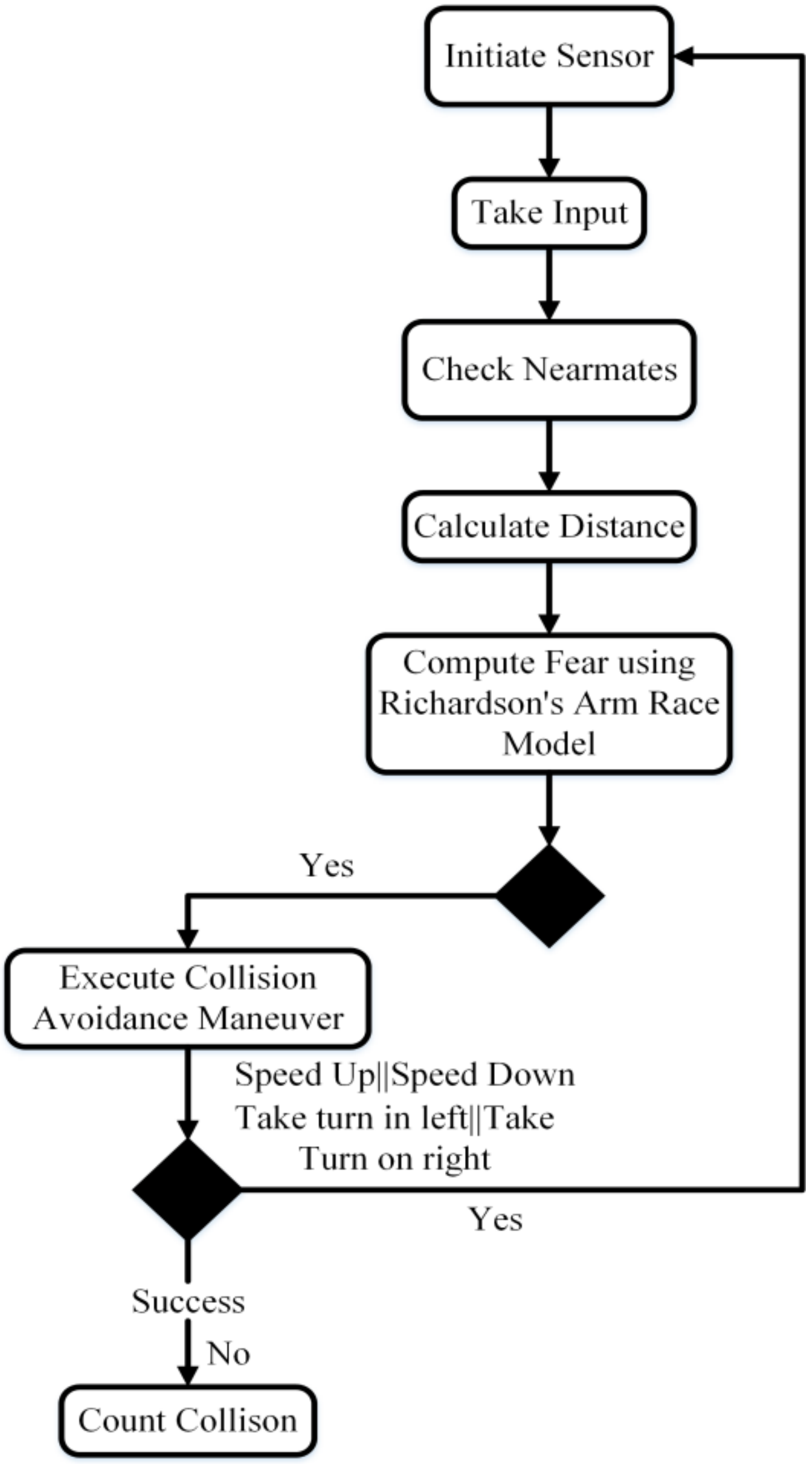

Fig. 9 Activity diagram of the Richardson's arms race model inspired Agent-based collision detection and avoidance scheme

## 7.2 Simulation Environment

The purpose of the research work is to introduce a social agent within the AV, having the capability of finding out the intentions of neighbouring AVs and avoiding the collisions. To simulate the concept of this agent-based system a standard agent-based simulation platform is the main requirement. For this purpose, Net logo 5.3 has been utilised which is a standard agent-based simulation environment. The Net Logo environment consists of patches, links, and turtles. Figure 10 presents the experimental environment along with input and output parameters. The left side of the simulation world contains input sliders and the right side is presenting the simulation world, executing the scenarios of AVs moving in a flock like topologies. It is important to mention here that the social agent installed in AVs have been designed with the help of Richardson's arms race model, which were basically proposed to avoid the wars between two nations. Hence the red and black colour of AVs is depicting two different types of nations according to the description of Richardson's arms race model.

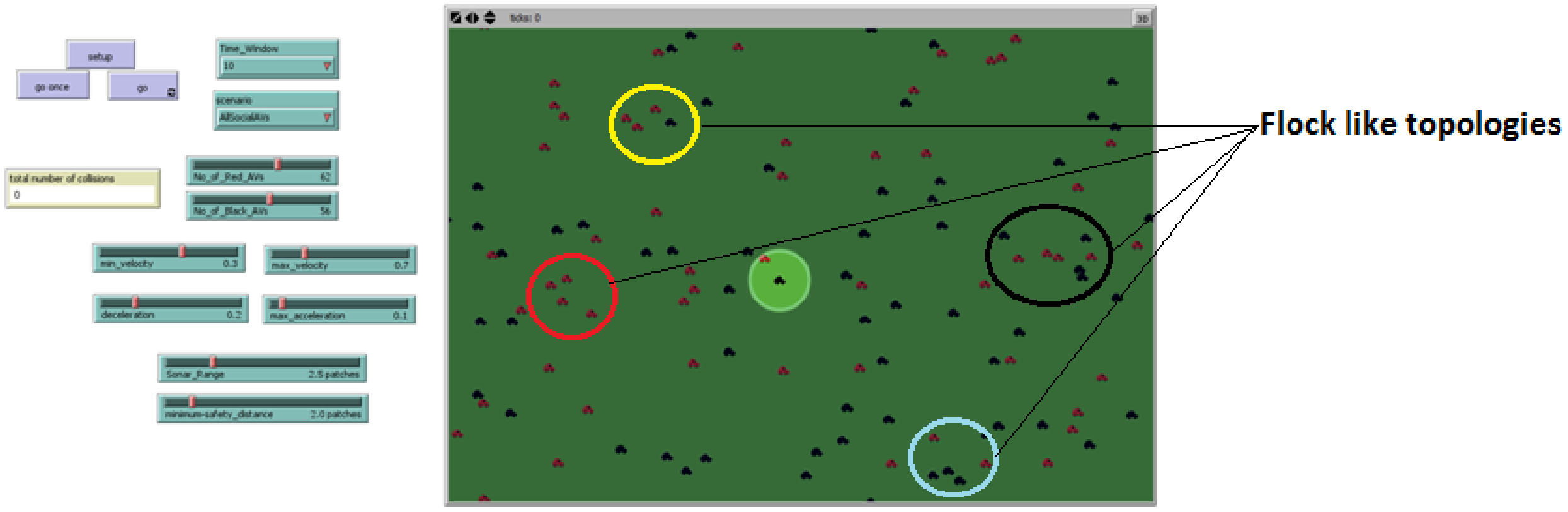

Fig. 10 Main Simulation Screen of Richardson's arms race model inspired Agent-based collision detection and avoidance scheme

### 7.2.1 Simulation Parameters

In this section, simulation parameters have been described. These simulation parameters can be seen in the form of input sliders in figure 10.

1. **NO of Red AVs**

We have used a slider "No of Red AVs" to set the no of red cars. We can set its value between 0 - 100.

2. **NO of Black AVs**

Similarly, a slider called “No of Black cars” controls the number of black cars. We can change the no of black cars by using this slider. We can set any value between 0 to 100.

**3. Scenarios:**

We used multiple scenarios in our simulation. These scenarios actually depict the simulation mode, i.e. social agent or random walk.

**4. Sonar Range**

This is also a slider representing the range of sonar. The sonar makes the AV able to detect the neighbouring vehicles. The higher the range of sonar the more the capability of AV in terms of generating the list of neighbouring AVs. The sonar range lies between 0 to 10 meters with the increment of 0.5 meters. However, for simulation purpose 1 patch of the simulation world is set to 1 meter.

**5. Min Velocity**

This slider is used to set the minimum velocity of the AVs. The slider ranges 0 to 0.5 with the increment of 0.1 m/s. No vehicle can decrease its speed less than the Min Velocity.

**6. Max Velocity**

Like “Min Velocity, there is also a slider, Max Velocity, to determine the maximum velocity of vehicles. Its value can vary from 0.6 to 1 with the increment of 0.1 m/s. The vehicles must not exceed this limit. This is the upper bound of both red and black AVs.

**7. Deceleration**

Deceleration slider helps in determining the deceleration rate of the Red and Black AVs. Its value lies between 0.1 to 0.5 $m/s^2$. This parameter can also be used to decrease the speed of any vehicle to a set value if the vehicle crosses the max velocity range.

**8. Max acceleration**

Max acceleration slider helps the Red and Black AVs to increase their acceleration rate. Its value lies between 0 to 0.1 $m/s^2$.

9. **Minimum Safety Distance**

This slider determines the minimum distance that each vehicle should maintain from the nearest AV. Its value lies in the range of 1.5 to 5 meters.

### 7.2.2 Simulation Experiment Design

In this section, the detailed experimental design has been proposed to test the performance of random walk based and Richardson's arms race model installed AVs in terms of collision avoidance. Four different experimental sets have been devised. The experimental set 1, presented in table 1, consists of 5 tests with same simulation parameters, but a different number of red and black AVs. These 5 tests of experimental set 1 help in testing the behaviour of AVs, in terms of collision avoidance for both random walk and Richardson's arms race model, with low speed, minimum acceleration rate, minimum deceleration rate, minimum safety distance and low sonar range.

TABLE 1
EXPERIMENT SET 1: PARAMETERS AND THEIR VALUES FOR40 RED AND 40 BLACK AVS ALONG WITH SIMULATION MODE

| Exp # | Number of Red_AVs | Number of Black_AVs | Min Velocity Range | Max Velocity Range | Min Acceleration Rate | Deceleration Rate | Min Safety Distance | Sonar Range | Simulation Mode |
|---|---|---|---|---|---|---|---|---|---|
| 1 | 40 | 40 | 0.3 | 0.3 | 0.1 | 0.1 | 1 | 2.5 | Random Walk/ Social Agent based |
| 2 | 50 | 50 | 0.3 | 0.3 | 0.1 | 0.1 | 1 | 2.5 | Random Walk/ Social Agent based |
| 3 | 60 | 60 | 0.3 | 0.3 | 0.1 | 0.1 | 1 | 2.5 | Random Walk/ Social Agent based |
| 4 | 70 | 70 | 0.3 | 0.3 | 0.1 | 0.1 | 1 | 2.5 | Random Walk/ Social Agent based |
| 5 | 80 | 80 | 0.3 | 0.3 | 0.1 | 0.1 | 1 | 2.5 | Random Walk/ Social Agent based |

6.

Table 2 presents the 5 tests of experimental set 2. This experimental set has been designed to measure the performance of proposed Richardson's arms race model based collision avoidance scheme with Random walk based collision avoidance scheme having the high-velocity range, i.e. 0.5 – 0.9 m/s. The remaining parameters have the same values as experimental set 1.

TABLE 2
EXPERIMENT SET 2: PARAMETERS AND THEIR VALUES FOR 40 RED AND 40 BLACK AVS ALONG WITH SIMULATION MODE

| Exp # | Number of Red_AVs | Number of Black_AVs | Min Velocity Range | Max Velocity Range | Min Acceleration Rate | Deceleration Rate | Min Safety Distance | Sonar Range | Simulation Mode |
|---|---|---|---|---|---|---|---|---|---|
| 1 | 40 | 40 | 0.5 | 0.9 | 0.1 | 0.3 | 1 | 2.5 | Random Walk/ Social Agent based |
| 2 | 50 | 50 | 0.5 | 0.9 | 0.1 | 0.3 | 1 | 2.5 | Random Walk/ Social Agent based |
| 3 | 60 | 60 | 0.5 | 0.9 | 0.1 | 0.3 | 1 | 2.5 | Random Walk/ Social Agent based |
| 4 | 70 | 70 | 0.5 | 0.9 | 0.1 | 0.3 | 1 | 2.5 | Random Walk/ Social Agent based |
| 5 | 80 | 80 | 0.5 | 0.9 | 0.1 | 0.3 | 1 | 2.5 | Random Walk/ Social Agent based |

# 8. Results and Discussion

In this section results of the above-mentioned experiments along with detailed discussion is presented.

## 8.1 Results and Discussion of Experiment Set 1

Table 3 to 7 presents the results of an experiment set 1 in terms of a mean number of collisions along their standard deviation values.  From the results of table 3, it can be seen that the proposed social agent based collision avoidance scheme outperforms random walk based collision avoidance scheme when the total number of AVs is 80. In first result, there are 1248 collisions, when the AVs follow the random walk pattern for travelling. However, using a social agent based technique the number of collisions decreased to the figure of 268.25. It means that the proposed technique helps AVs to avoid the collisions by having the know-how of each other's current position using Sensory and Artificial Thalamus module and mirroring module.  In the same way, the 6th result of table 3 shows that using the proposed technique, there are only 270 collisions as compared to the 1166.12 collisions in the case of Random walk based movement of AVs. From the experiment set 1, presented in table 1, it can be seen that the remaining simulation parameters are same, but only the number of AVs is varied. Table 4 presents a number of collisions for 50 red and 50 black AVs. In comparison with table 3, it can be seen that the number of collisions has been increased for a higher number of vehicles. The first entry in table 4 shows that there are 1883.25 collisions if AVs follow the random walk. Whereas there are only 1248.12, the first entry of table 3, collisions using the random walk. However, there is a slight increase in the number of collisions using a social agent model in table 4 as compared to the table 3. It means that the increase in a number of vehicles leads to the increase in the number of collisions, but still far less than the random walk. If we compare the results within table 4 then it can be seen that social agent enabled AVs have a low number of accidents as compared to the random walk based AVs. From the remaining tables (5 - 7), it can be seen that the social agent enabled AVs outperform random walk based AVs in terms of fewer collisions. An interesting fact is that for random walk based AVs, a number of collisions increase with an increase in the number of AVs, whereas for proposed technique, the range of collisions does not cross the figure of 300 collisions. Figures 11 through 15 are the graphical representations of tables 3 through 7 respectively.

TABLE 3

RESULTS OF EXPERIMENT SET 1 IN TERMS OF MEAN NUMBER OF COLLISIONS ALONG WITH STDEV FOR RANDOM WALK AND SOCIAL AGENT BASED COLLISION AVOIDANCE TECHNIQUE WITH 40 RED AND 40 BLACK AVS

| Random Walk Based | | Social Agent based | |
|---|---|---|---|
| Mean | Stdev | Mean | Stdev |
| 1248.12 | 3.56 | 268.25 | 1.66 |
| 1347 | 5.95 | 268.25 | 1.28 |
| 1488.12 | 10.9 | 277.75 | 1.28 |
| 1080 | 6.5 | 270.25 | 2.12 |
| 1433 | 1.41 | 263.5 | 1.51 |
| 1166.12 | 8.33 | 270.37 | 0.74 |

TABLE 4

RESULTS OF EXPERIMENT SET 1 IN TERMS OF MEAN NUMBER OF COLLISIONS ALONG WITH STDEV FOR RANDOM WALK AND SOCIAL AGENT BASED COLLISION AVOIDANCE TECHNIQUE WITH 50 RED AND 50 BLACK AVS

| Random Walk Based | | Social Agent based | |
|---|---|---|---|
| Mean | Stdev | Mean | Stdev |
| 1883.25 | 8.843884 | 267.12 | 1.457738 |
| 1977.25 | 10.20854 | 281.37 | 1.30247 |
| 2087.75 | 17.18596 | 277.37 | 0.744024 |
| 1955 | 13.0384 | 271.75 | 0.707107 |
| 1986.37 | 6.162965 | 281.75 | 0.707107 |
| 1961 | 10.69045 | 277.37 | 1.30247 |

TABLE 5

RESULTS OF EXPERIMENT SET 1 IN TERMS OF MEAN NUMBER OF COLLISIONS ALONG WITH STDEV FOR RANDOM WALK AND SOCIAL AGENT BASED COLLISION AVOIDANCE TECHNIQUE WITH 60 RED AND 60 BLACK AVS

| Random Walk Based | | Social Agent based | |
|---|---|---|---|
| Mean | Stdev | Mean | Stdev |
| 3486.62 | 18.41535 | 271.25 | 1.28174 |
| 3191.62 | 10.15505 | 285.5 | 1.511858 |
| 2750.37 | 21.15209 | 276.5 | 0.534522 |
| 3066.37 | 10.92752 | 273.87 | 1.642081 |
| 3326.37 | 23.56715 | 264.62 | 0.916125 |
| 3291.75 | 14.53813 | 273.37 | 1.30247 |

TABLE 6

RESULTS OF EXPERIMENT SET 1 IN TERMS OF MEAN NUMBER OF COLLISIONS ALONG WITH STDEV FOR RANDOM WALK AND SOCIAL AGENT BASED COLLISION AVOIDANCE TECHNIQUE WITH 70 RED AND 70 BLACK AVS

| Random Walk Based | | Social Agent based | |
|---|---|---|---|
| Mean | Stdev | Mean | Stdev |
| 3935 | 20.79148 | 291.25 | 1.38873 |
| 3366.37 | 18.43086 | 256.5 | 0.92582 |
| 4395.87 | 21.33031 | 288.37 | 1.06066 |
| 5052 | 20.73644 | 251.37 | 0.916125 |
| 4000.87 | 26.12299 | 267.25 | 1.832251 |
| 3964 | 29.55866 | 282 | 1.195229 |

TABLE 7

RESULTS OF EXPERIMENT SET 1 IN TERMS OF MEAN NUMBER OF COLLISIONS ALONG WITH STDEV FOR RANDOM WALK AND SOCIAL AGENT BASED COLLISION AVOIDANCE TECHNIQUE WITH 80 RED AND 80 BLACK AVS

| Random Walk Based | | Social Agent based | |
|---|---|---|---|
| Mean | Stdev | Mean | Stdev |
| 5741 | 20.79835 | 265.25 | 2.12132 |
| 5205.5 | 26.75284 | 268.62 | 1.59799 |
| 5781.62 | 26.90161 | 260 | 1.511858 |
| 5404.62 | 28.71504 | 275.25 | 1.164965 |
| 5644.62 | 26.47337 | 281.25 | 1.035098 |
| 4944.25 | 18.82817 | 282.5 | 1.414214 |

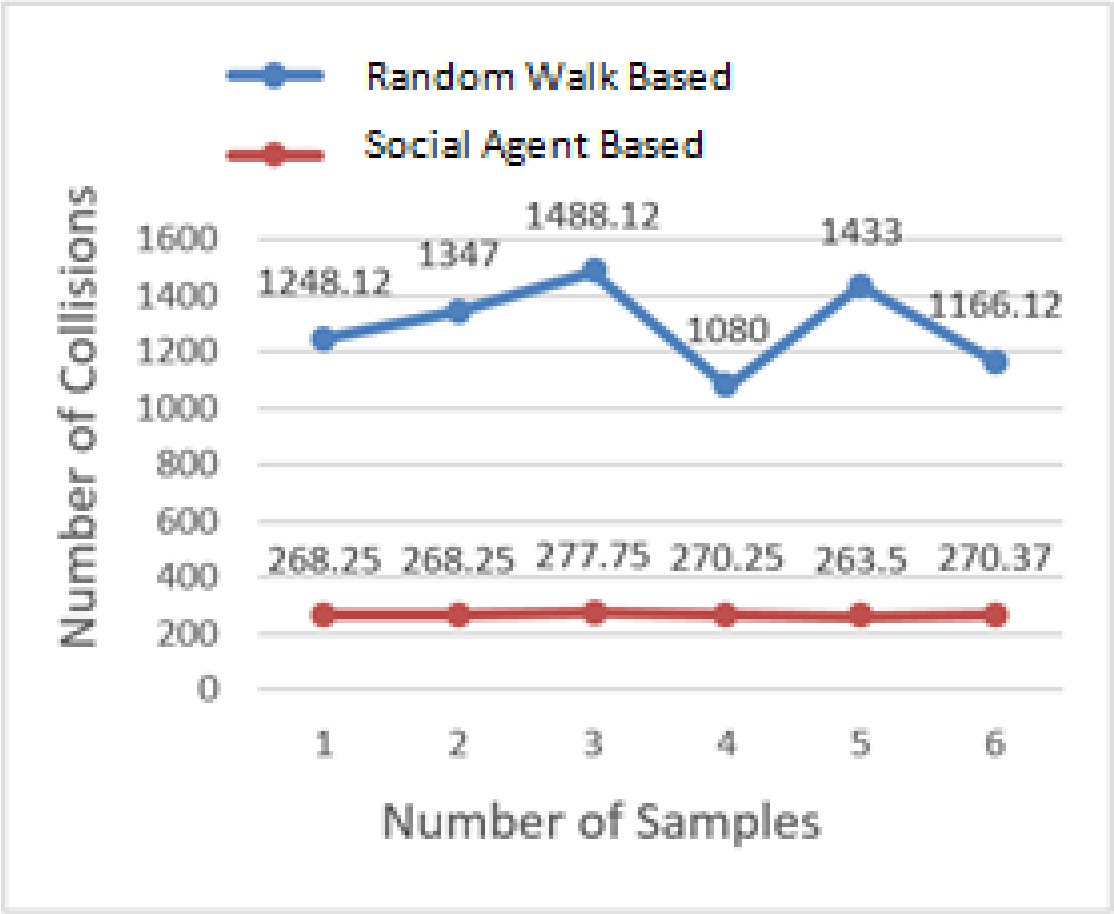

Fig. 11 Graphical representation of the results of table 3

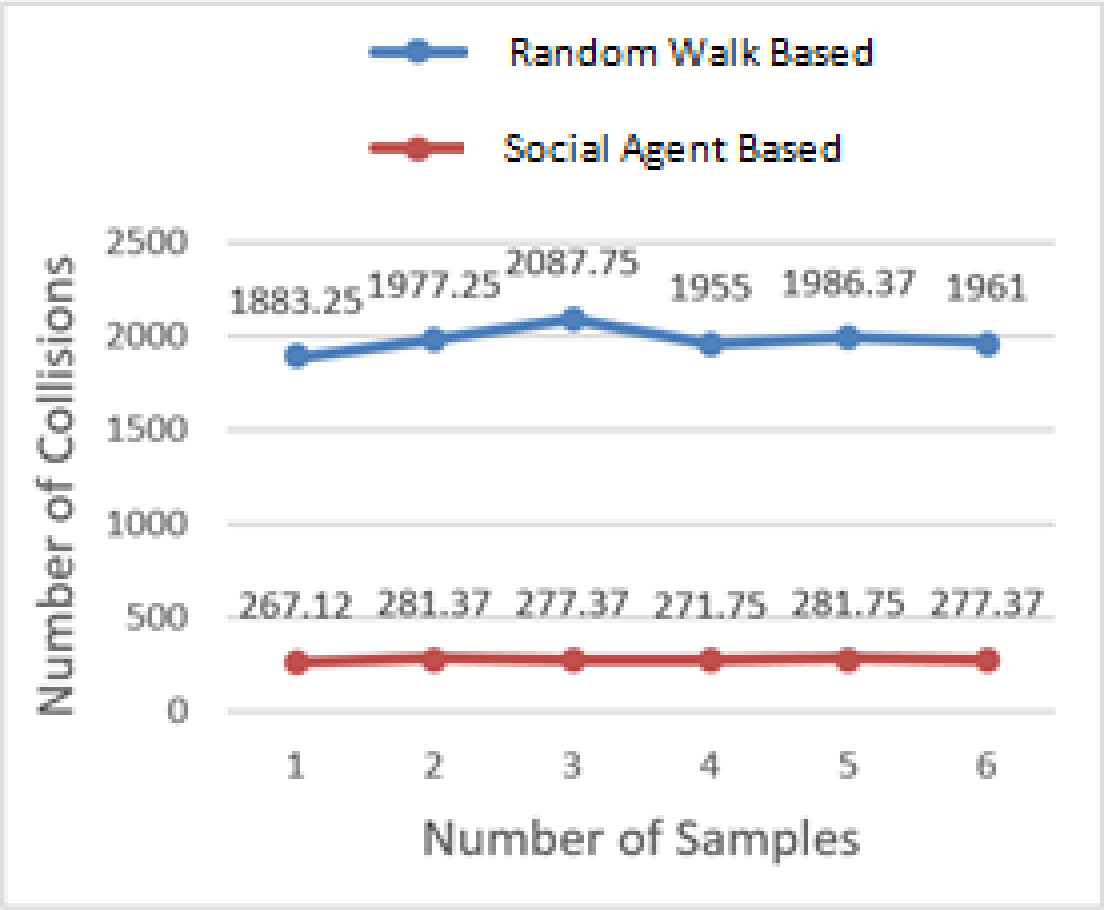

Fig. 12 Graphical representation of the results of table 4

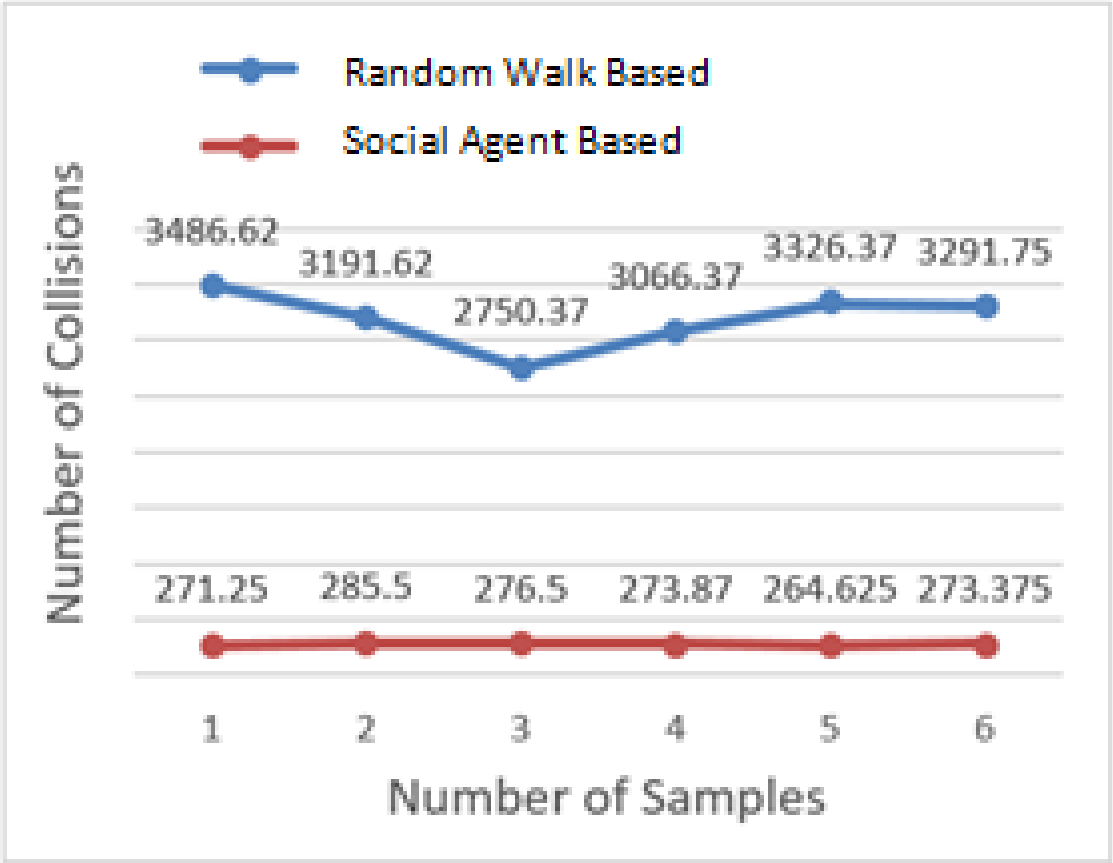

Fig. 13 Graphical representation of the results of table 5

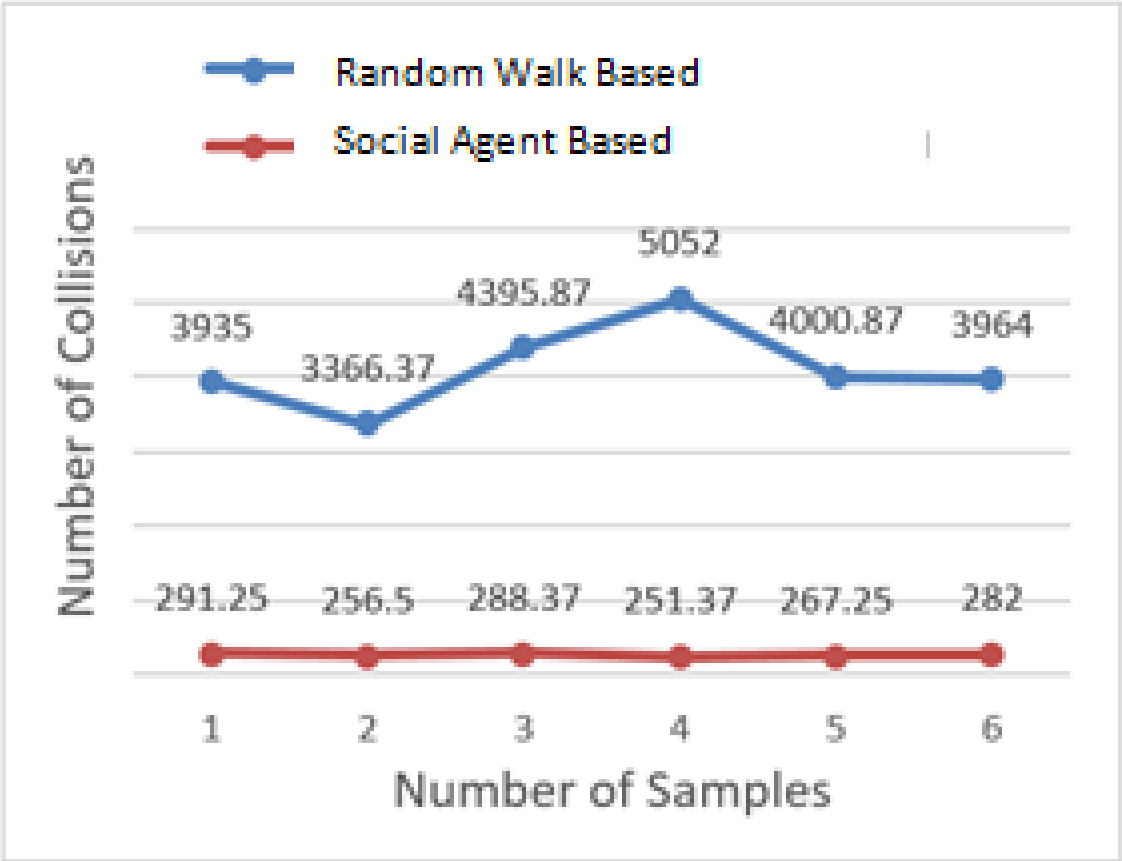

Fig. 14 Graphical representation of the results of table 6

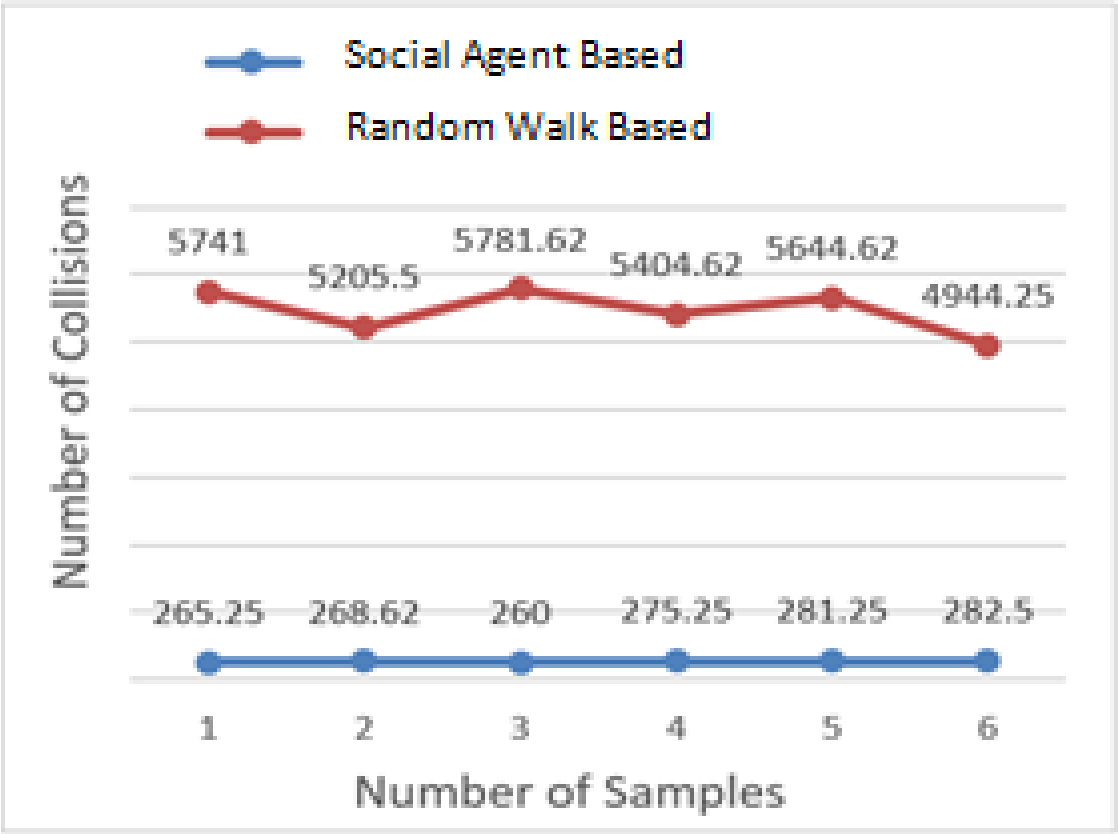

Fig. 15 Graphical representation of the results of table 7

## 8.2 Results and Discussion of Experiment Set 2

Table 8 to 12 presents the simulation results of the experiment set 2. If we recall experiment set 2, table 2, then it is different from the experiment set 1 in terms of minimum and maximum velocity range. Now the range has been set between 0.5 to 0.9. Whereas, the deceleration rate is same, i.e. 0.1 $m/s^2$. In these experiments, the performance of social agent enabled AVs is compared with random walk based AVs in terms of collision avoidance. Table 8 presents the comparison of both techniques for experiment set 2 having 40 red and 40 black AVs. In the first result, there are 1251.75 collisions when AVs use the Random walk pattern to travel. Whereas the number of collisions have been minimized using the social agent model and there are only 507.5 collisions. In the same way, the other entries of table 8 are 1312.5, 1422.62, 1348.75, 1320.75, and 1499.75 for random walk based technique. Whereas for the Richardson's arms race model technique these numbers of collisions are 496.75, 456.87, 487.5, 440.62, and 542.25 respectively. If we compare these results, then it can be seen that our proposed scheme have performed fewer collisions as compared to the Random walk based AVs. The analysis of remaining tables 9 to 12 proves that social agent based collision avoidance scheme outperforms Random walk based collision avoidance scheme. Figures 16 to 20 are the graphical representations of tables 9 to 12 respectively.

TABLE 8

RESULTS OF EXPERIMENT SET 2 IN TERMS OF MEAN NUMBER OF COLLISIONS ALONG WITH STDEV FOR RANDOM WALK AND SOCIAL AGENT BASED COLLISION AVOIDANCE TECHNIQUE WITH 40 RED AND 40 BLACK AVS

| Random Walk Based | | Social Agent based | |
|---|---|---|---|
| **Mean** | **Stdev** | **Mean** | **Stdev** |
| 1251.75 | 6.943651 | 507.5 | 2.203893 |
| 1312.5 | 6.88684 | 496.75 | 1.908627 |
| 1422.62 | 2.722263 | 456.87 | 1.356203 |
| 1348.75 | 6.692213 | 487.5 | 3.585686 |
| 1320.75 | 4.682795 | 440.62 | 2.615203 |
| 1499.75 | 6.541079 | 542.25 | 2.712405 |

TABLE 9

RESULTS OF EXPERIMENT SET 2 IN TERMS OF MEAN NUMBER OF COLLISIONS ALONG WITH STDEV FOR RANDOM WALK AND SOCIAL AGENT BASED COLLISION AVOIDANCE TECHNIQUE WITH 50 RED AND 50 BLACK AVS

| Random Walk Based | | Social Agent based | |
|---|---|---|---|
| **Mean** | **Stdev** | **Mean** | **Stdev** |
| 2143.75 | 6.798109 | 980.62 | 4.373214 |
| 2227.75 | 7.421013 | 984.25 | 6.453128 |
| 2192.5 | 10.47446 | 976.62 | 5.730557 |
| 2206.12 | 8.741322 | 970.87 | 4.853202 |
| 2252.37 | 15.21219 | 960.25 | 4.399675 |
| 2177.25 | 8.972179 | 981.62 | 4.501984 |

TABLE 10

RESULTS OF EXPERIMENT SET 2 IN TERMS OF MEAN NUMBER OF COLLISIONS ALONG WITH STDEV FOR RANDOM WALK AND SOCIAL AGENT BASED COLLISION AVOIDANCE TECHNIQUE WITH 60 RED AND 60 BLACK AVS

| Random Walk Based | | Social Agent based | |
|---|---|---|---|
| **Mean** | **Stdev** | **Mean** | **Stdev** |
| 2951 | 11.3641 | 1034.25 | 5.725881 |
| 3003 | 14.67749 | 1055.87 | 5.591767 |
| 2845.37 | 15.91888 | 1100.62 | 4.033343 |
| 3042.87 | 18.52749 | 1131.5 | 6.41427 |
| 3217.12 | 12.11183 | 1083.87 | 5.890367 |
| 3102.75 | 17.20257 | 1039.62 | 3.20435 |

TABLE 11

RESULTS OF EXPERIMENT SET 2 IN TERMS OF MEAN NUMBER OF COLLISIONS ALONG WITH STDEV FOR RANDOM WALK AND SOCIAL AGENT BASED COLLISION AVOIDANCE TECHNIQUE WITH 70 RED AND 70 BLACK AVS

| Random Walk Based | | Social Agent based | |
|---|---|---|---|
| **Mean** | **Stdev** | **Mean** | **Stdev** |
| 4090.75 | 17.26888 | 1584.12 | 8.542959 |
| 4123.37 | 18.89019 | 1587 | 8.815571 |
| 4002.5 | 15.29706 | 1651.12 | 8.131728 |
| 4129.87 | 14.26722 | 1641.25 | 9.346504 |
| 4286.5 | 20.25551 | 1611.37 | 5.069164 |
| 3995.87 | 20.90412 | 1562 | 6.436503 |

TABLE 12

RESULTS OF EXPERIMENT SET 2 IN TERMS OF MEAN NUMBER OF COLLISIONS ALONG WITH STDEV FOR RANDOM WALK AND SOCIAL AGENT BASED COLLISION AVOIDANCE TECHNIQUE WITH 80 RED AND 80 BLACK AVS

| Random Walk Based | | Social Agent based | |
|---|---|---|---|
| **Mean** | **Stdev** | **Mean** | **Stdev** |
| 5254.75 | 24.76028 | 1281.5 | 5.318432 |
| 5189.25 | 21.90727 | 1355.75 | 7.554563 |
| 5629.37 | 22.36667 | 1325.5 | 6.524678 |
| 5199.37 | 25.36554 | 1304.12 | 8.025629 |
| 5320.87 | 21.6428 | 1129.12 | 6.685539 |
| 5562.87 | 18.90153 | 1039.62 | 4.068608 |

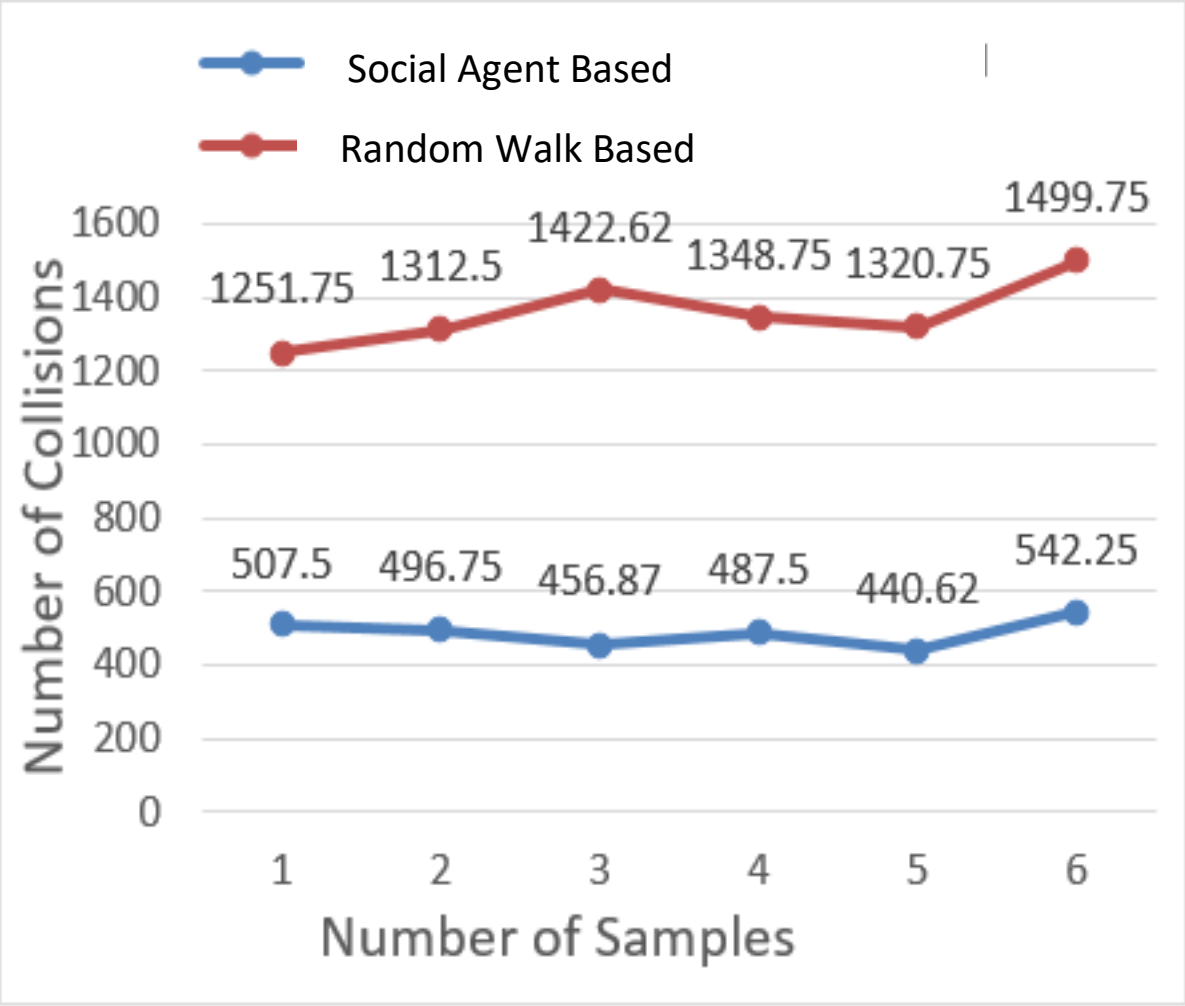

Fig. 16 Graphical representation of the results of table 8

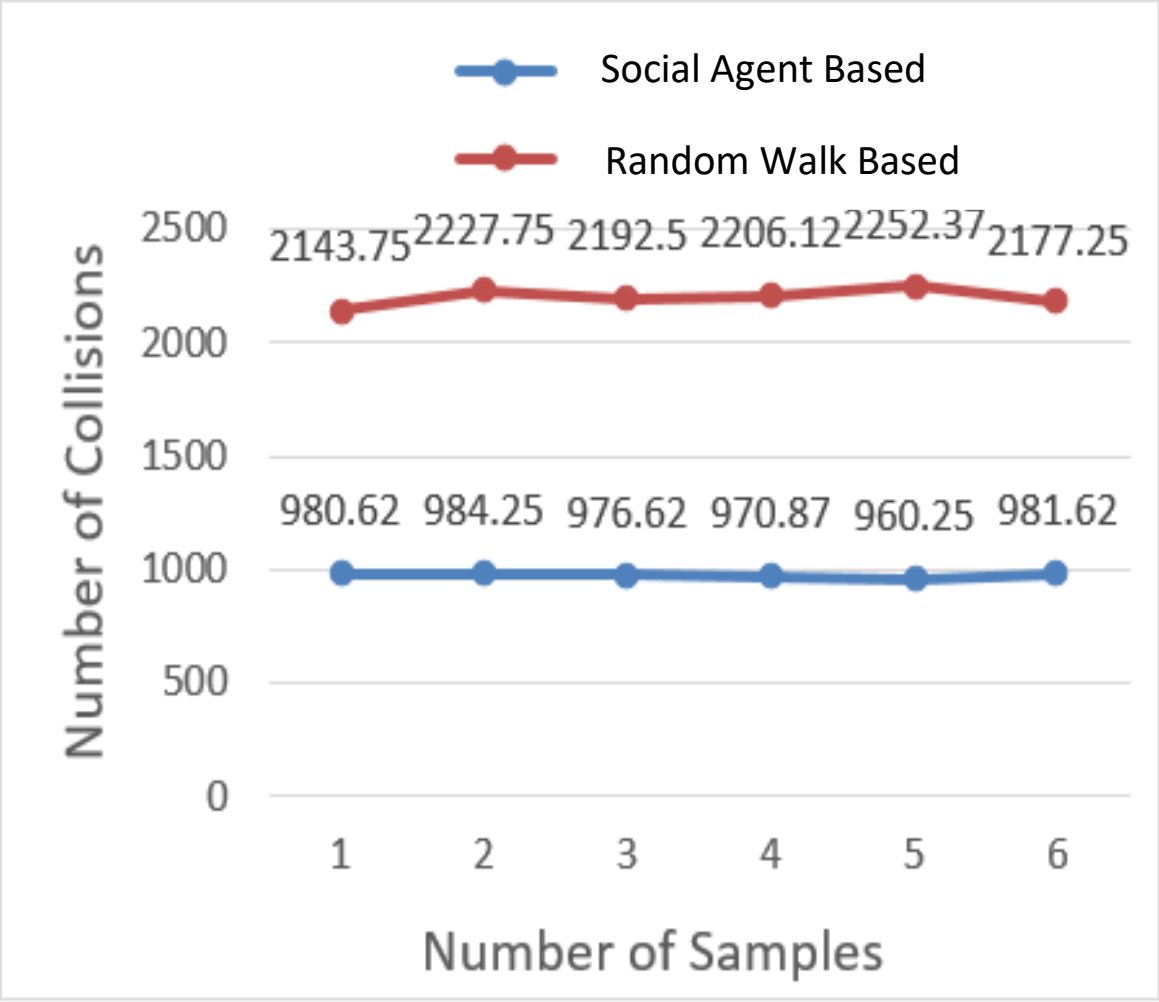

Fig. 17 Graphical representation of the results of table 9

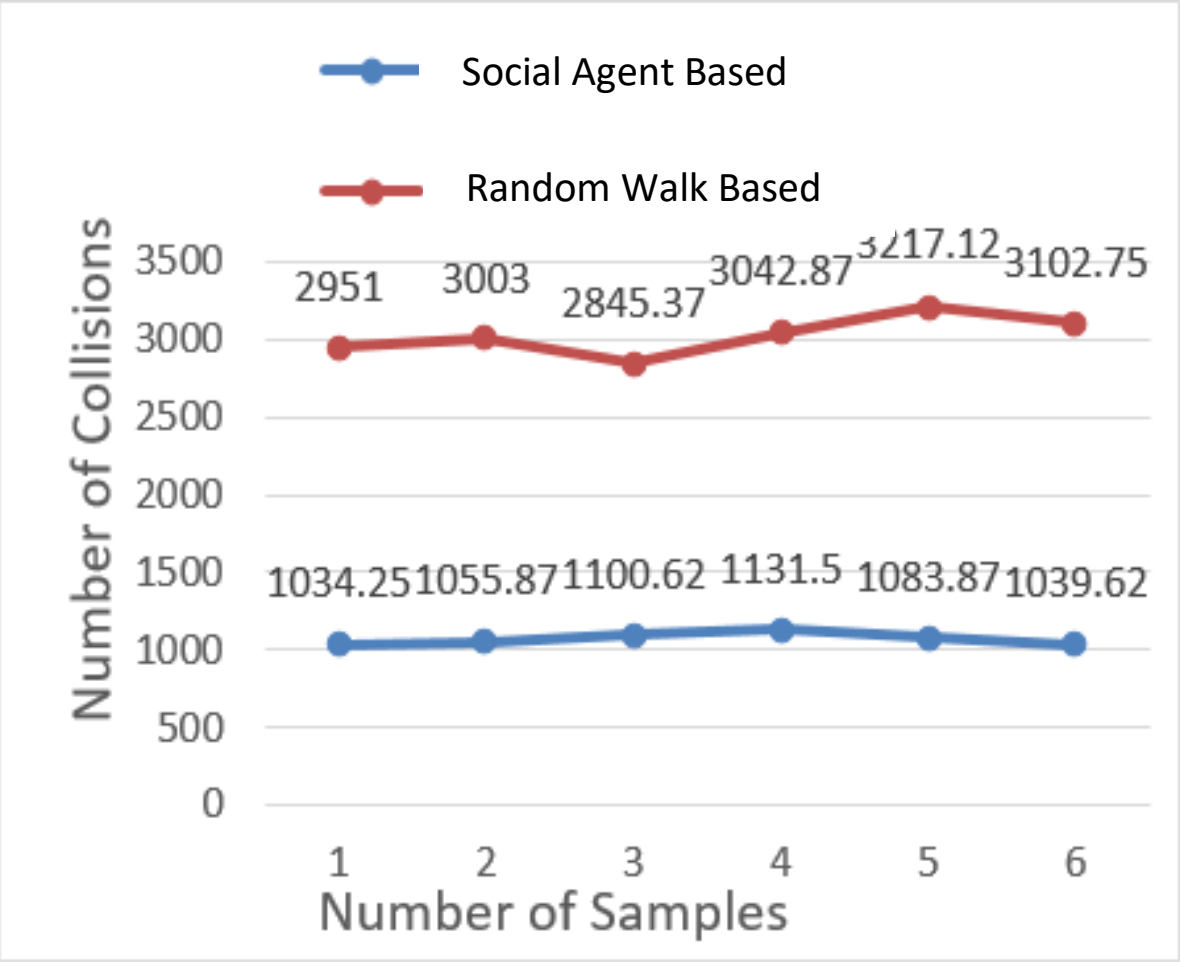

Fig. 18 Graphical representation of the results of table 10

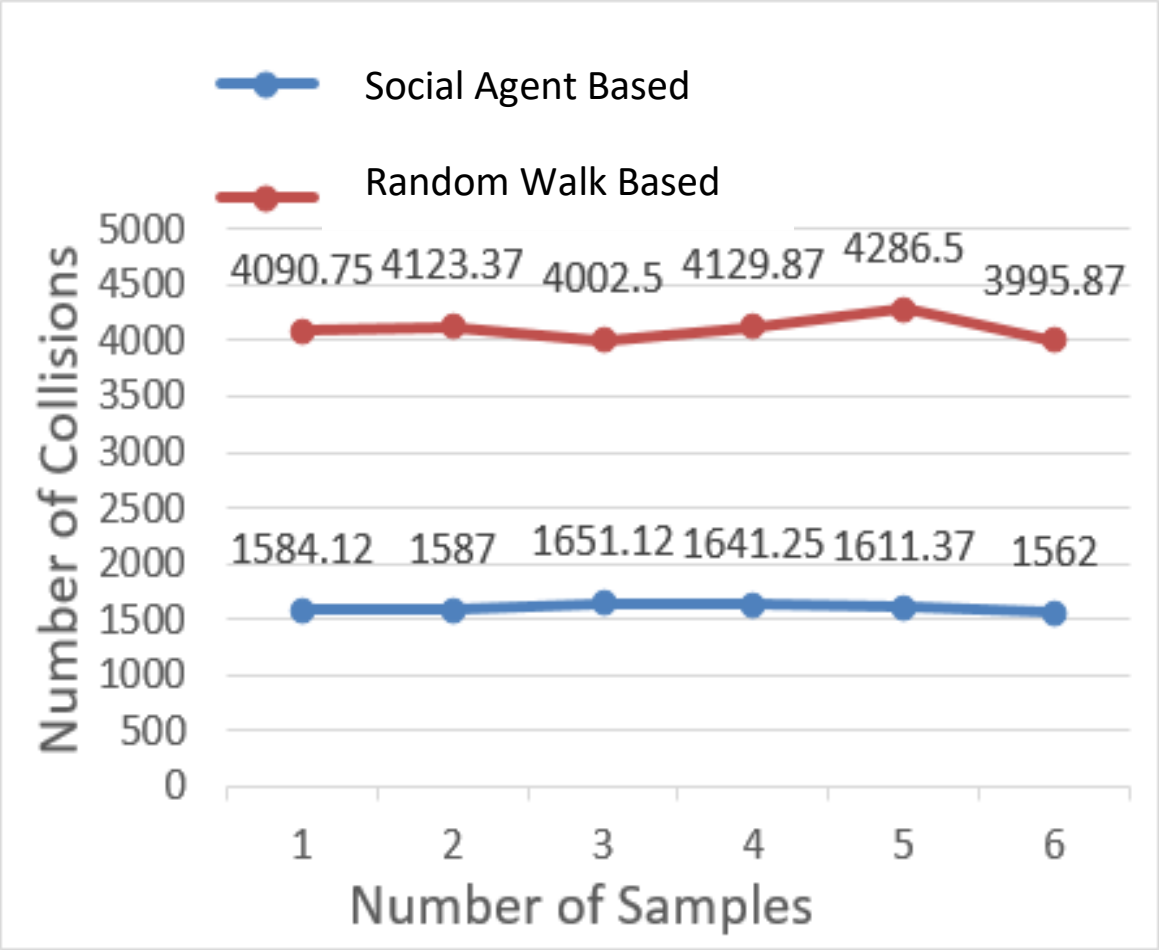

Fig. 19 Graphical representation of the results of table 11

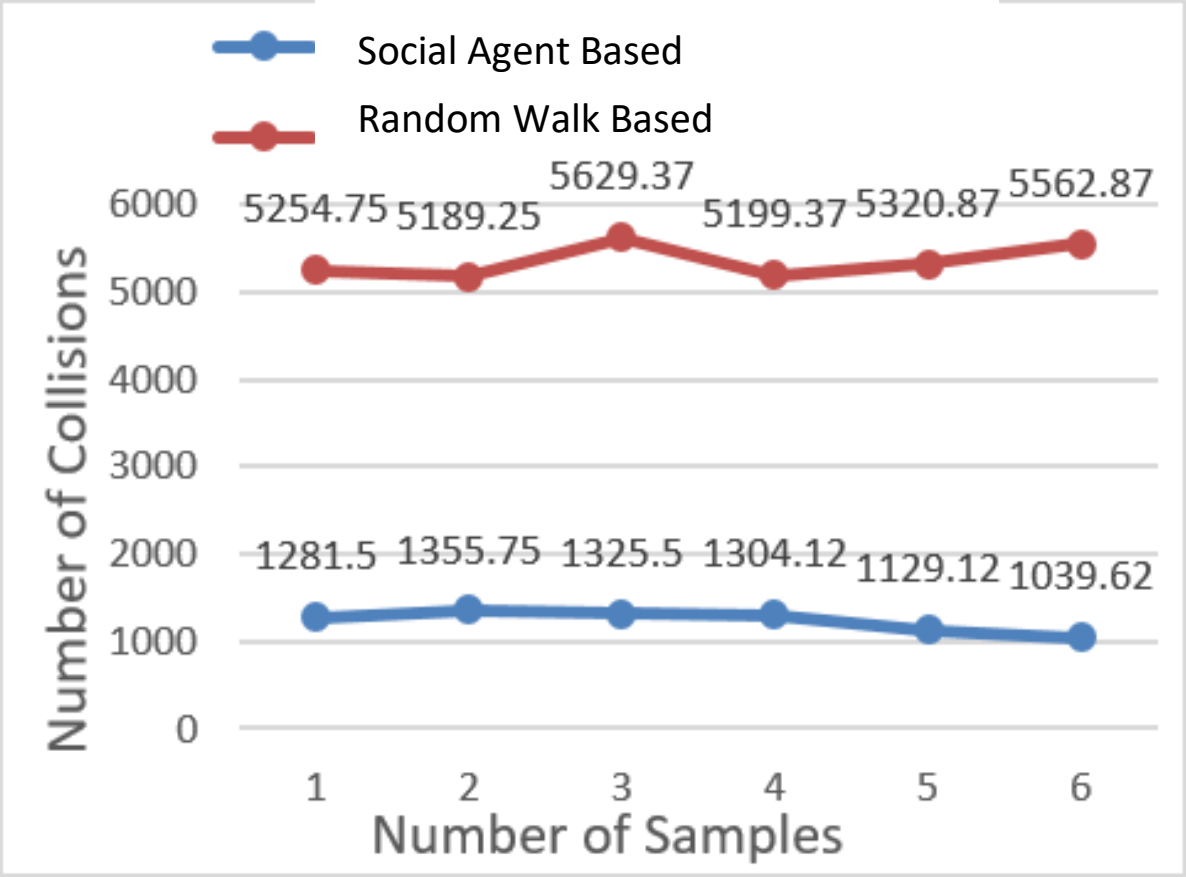

Fig .20 Graphical representation of the results of table 12

Now if we compare the results of the experiment set 1 and set 2, then it can be seen that Richardson's arms race model with high velocity has a higher number of collisions as compared to the Richardson's arms race model with low velocity. Let us compare the table 4, set 1 for 50 red and 50 black AVs, with the table 9 of set 2 for 50 red and 50 black AVs. The first result of table 4 shows that there are 267.12 collisions, whereas the first entry of table 9 shows 980.62 collisions. Furthermore, the 3rd entry of table 4 shows 277.37 collisions whereas there are only 976.62 collisions as shown in the 3rd entry in table 9. If we keep performing the analysis of other tables of experiment sets, 1 and 2, then it can be observed very clearly that Richardson's arms race model with low velocity and low deceleration rate can outperform the Richardson's arms race model based collision avoidance technique with high-velocity and low deceleration rate.

## 9. Practical validation of the Proposed Social Agent Functionality

To give the proof of concept and to perform the rigorous validation of the proposed social agent, we have performed field tests. For this purpose, a prototype AV platform has been built, which is equipped with sonar sensors and Arduino microcontroller. Furthermore, the functionality of the proposed social agent has been coded using the Integrated Development Environment of Arduino Microcontroller (IDEAM). Figure 21 presents the field experiment, which is performed with three human-driven motorcycles and specially built AV installed with a social agent.

### Infield Experiment Design using Flock Like Topology

(i) Three human-driven motorcycles maneuvering around the prototype AV platform. The leading motorcyclist drive with different acceleration and deceleration rate . Whereas, the motorcyclists driving on both lateral sides drive with the same speed of AV and increase and decrease their lateral distance from AV in a random fashion.

(ii) The results of each test have been traced into a log file every millisecond.

**Results and Discussion:** Table 13 presents the results of in-field experiments. Total 8 tests have been performed to validate the performance of the social agent. If we study the results of the first test then it can be seen that social AV takes 0.00138 seconds to sense the three neighbouring vehicles and found the front vehicle at the distance of 2.6 ft, and Lateral Left (LL) and Lateral Right (LR) vehicles in 1.8 and 3.2 ft respectively. In next step social agent takes 0.000002 seconds to compute the nearest vehicle and declared LL the nearest one. During the experiment, when the LL drifted towards AV and reached the preset safety threshold the mirroring module of social agent copied the drifting angle of LL and executed turn left manoeuvre in 0.000008 seconds. The total time taken by a social agent from sensing the neighbours to execute the collision avoidance manoeuvre is 0.001408 seconds. In the same way, the other tests prove the effectiveness of the proposed approach regarding collision avoidance in a very short time.

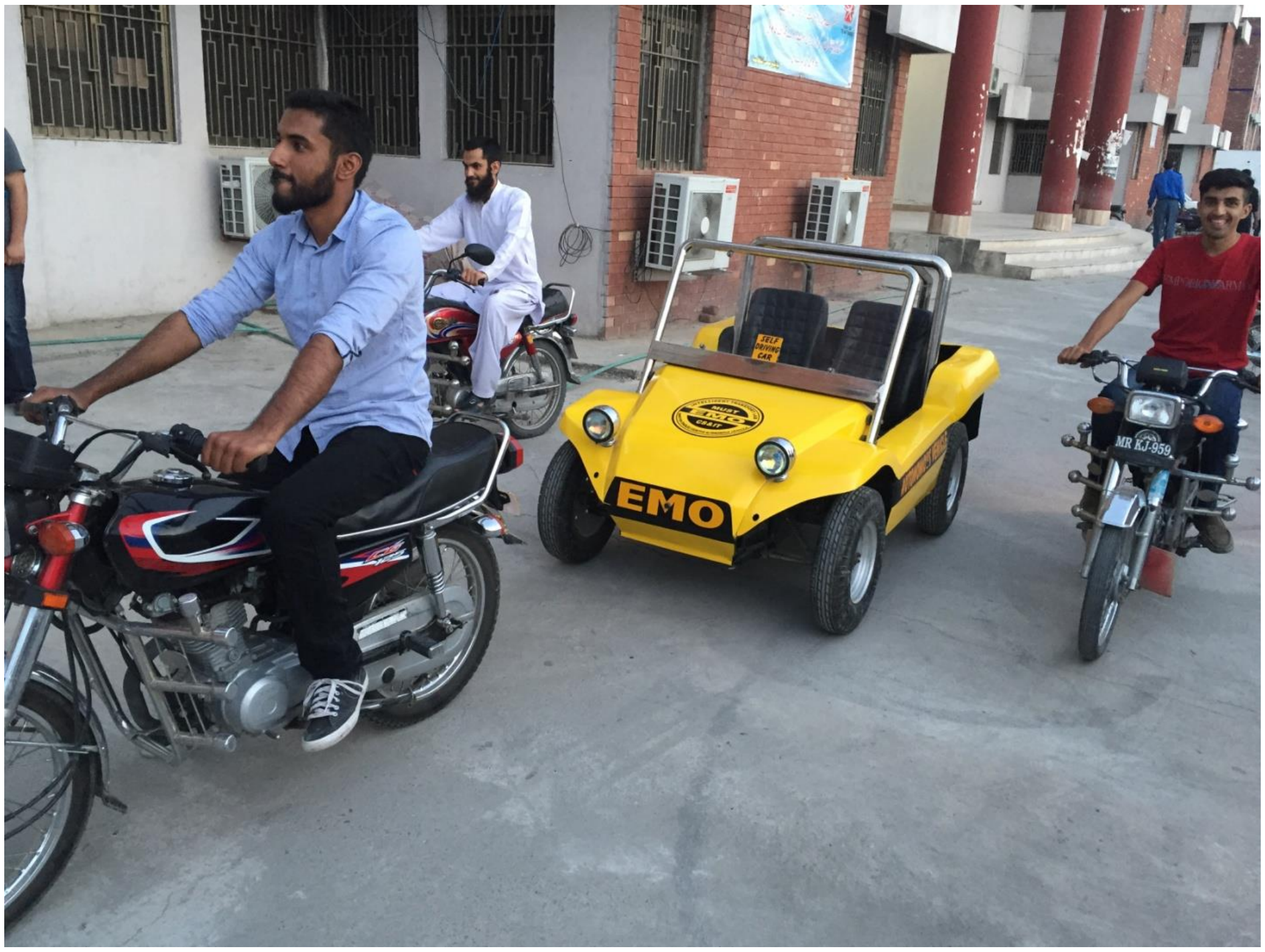

Fig. 21 Infield experiment using Flock Like Topology

TABLE 13

RESULTS OF INFIELD EXPERIMENTS IN TERMS OF TIME TAKEN BY THE SOCIAL AGENT FOR THE COLLISION AVOIDANCE in the flock like topology

Performance Metrices

| Total Time (sec) | 0.001408 | 0.001436 | 0.001424 | 0.001408 | 0.00144 | 0.001408 | 0.001424 | 0.001436 | Average Total Time (sec) |
|---|---|---|---|---|---|---|---|---|---|
| Decision Execution Time (sec) | 0.000008 | 0.000036 | 0.000016 | 0.000032 | 0.000032 | 0.000032 | 0.000016 | 0.000036 | |
| Action taken by Social Agent | Turn Left | Brake Applied | Turn Right | Turn Left | Brake Applied | Turn Left | Turn Right | Brake Applied | |
| Nearest Vehicle | Left Lateral | Front | Right Lateral | Left Lateral | Front | Left Lateral | Right Lateral | Front | |
| Computing Nearest Neighbor Time (sec) | 0.00002 | 0.00002 | 0.000024 | 0.000032 | 0.000032 | 0.000032 | 0.000024 | 0.00002 | 0.001423 |
| Neighboring Vehicles Positions (ft) (axis 1–6) | .LR .F .LL | .LR .LL .F | .F .LL .LR | .F .LR .LL | .LR .LL .F | .LR .F .LL | .LL .F .LR | .LR .LL .F | |
| Sensing Time (sec) | 0.00138 | 0.00138 | 0.001384 | 0.001344 | 0.001376 | 0.001344 | 0.001384 | 0.00138 | |

Timeline

## 10. Comparison With The Existing State Of The Art

As discusses in section 2, Motivation behind research work, we have not found any research work to addresses the presented problem. However, we have found a mirror inspired cooperative perception based collision avoidance scheme by Kim and Liu (33) which is close to our proposed research in a single aspect. Kim and Liu (33) utilised the concept of mirror neurons to propose the longitudinal and lateral motion control mechanism using cooperative perception. The presented model is a macroscopic model, which takes into account the overall behaviour of the AVs. Though the authors have claimed to use the human mirror Neurons to guess the intention of leading vehicles but it relies on cooperative perception. However, the intention aware mechanism regarding laterally moving vehicles has not been devised that help the AVs to optimise their latitude control and help them in avoiding lateral collisions. Furthermore, the cooperative perception has been utilised, which depends on the wireless medium. According to (33), cooperative perception is suitable in making short-term perspective driving the decision for hidden collision Avoidance but it doesn't help in defining the longitudinal and lateral control mechanism, which helps the autonomous vehicles to avoid the collisions from the non-hidden neighbouring vehicles, travelling in side by side fashion. Furthermore, the cooperative perception between AVs has been supposed to be made using Wireless access for Vehicular Environment (WAVE) as a communication medium. However, WAVE is proved to be a non-suitable solution for sharing local information. According to (34), WAVE has not been found suitable to provide the reliable communication medium for increasing number of vehicles competing for the same channel within the same area. The real-time applications like road safety using cooperative perception requires less than 200 milliseconds delay (35) but it has been noted by (36) that due to data contention in the control channel of WAVE, data packets has to be resent many times and as a result the safety message delivery time exceeds 1000 milliseconds. Furthermore, (37) also noted that WAVE has been found not a suitable protocol for a periodic communication between vehicles to exchange the safety specific data. To measure the performance of IEEE 802.11n based Mirror Neuron Inspired Intention Awareness and Cooperative Perception Approach, we setup an experiment environment. The experiment platform consists of two toy AVs equipped with Arduino microcontrollers, GPS and wireless transceiver. To measure the performance of IEEE 802.11n based intention aware scheme, following metrics has been considered. ***Packet preparation time by sending Vehicle***, ***Average packet delay time between two vehicles, packet interpretation time by destination vehicle***, and ***Reaction time to avoid the collision***. The test results are presented in table 14. From the first test result, it can be seen that the sending vehicle takes 0.600372 seconds to prepare the message packet and then forward it to the destination vehicle. The message packet reaches to the destination vehicle with the delay of 0.152 seconds. After receiving the packet the destination vehicle takes 0.402408 seconds to understand the message hidden in the packet. In next step, the destination vehicle executes the collision avoidance manoeuvre in 0.000008 seconds. In this way, the total time taken by destination vehicle to avoid the collision is 1.154 seconds. We performed total eight experiments and it has been revealed that IEEE 802.11n based mirror neuron scheme takes 1.1109 seconds on average to avoid the collisions.

TABLE 14

Results of prtototype experiments in terms of the time taken by the IEEE 802.11n based mirror neuron inspired intention awareness and cooperative perception Approach (33) for the collision avoidance in the flock like topology

| Total Time (sec) | 1.154788 | 1.17464 | 1.078712 | 1.037716 | 1.266761 | 1.019812 | 1.080626 | 1.074752 | Average Total Time (sec) |
|---|---|---|---|---|---|---|---|---|---|
| Action Taken Time (sec) | 0.000008 | 0.000036 | 0.000016 | 0.000032 | 0.000032 | 0.000032 | 0.000016 | 0.000036 | 1.110975875 |
| Packet Interpretation Time (sec) | 0.402408 | 0.40224 | 0.40034 | 0.402328 | 0.600365 | 0.402418 | 0.402245 | 0.40034 | |
| Packet Transmission Delay Time (sec) | 0.152 | 0.172 | 0.078 | 0.035 | 0.066 | 0.017 | 0.078 | 0.074 | |
| Packet Preparation Time (sec) | 0.600372 | 0.600364 | 0.600356 | 0.600356 | 0.600364 | 0.600362 | 0.600365 | 0.600376 | |

In contrast to this research, we presented the microscopic model of collision avoidance using mentalizing and mirroring neuron without relying on cooperative perception. In conclusion, the proposed social agent based AVs can avoid rear end and lateral collisions in a flock like topology in 0.001423 seconds as compared to wireless based intention awareness system which takes 1.154 seconds for the same purpose. Hence the proposed scheme can avoid rear end and lateral collisions, in flock like topology, with the efficiency of 99.876 % as compared to the IEEE 802.11n based existing state of the art [10] mirroring neuron based collision avoidance scheme.

## 11. Conclusion

Artificial intelligence is the name of building machines, which act like human beings, by studying human beings. Autonomous vehicles are in town and no one can negate their importance. However, building collision free AVs is a challenging task. To address this, we proposed the concept of social AVs, which use the social interaction mechanism of human beings to avoid the potential collisions. Humans have special brain circuits that make them social and help them to interpret the intentions of other human beings and adapting the strategies to avoid the clashes. Inspired from this, we have proposed a concept of a social agent that help the AVs to avoid the collisions. In addition, a mathematical model inspired by Richardson's arms race model is proposed to emulate the social functions of human brain like mentalizing and mirroring. The performance of the proposed social agent is compared, using extensive experiment sets, with Random walk based collision avoidance strategy and it has been found that the proposed social agent based collision avoidance strategy is 78.52 % efficient than random walk based collision avoidance strategy and the practical validation results confirm that the proposed scheme can avoid rear end and lateral collisions with the efficiency of 99.876 % as compared to the IEEE 802.11n based existing state-of-the-art research work. Furthermore, the simulation results have provided optimal parameters, like optimal sonar range and different optimal speeds suitable for avoiding the road collisions in different road traffic situations. This research might be suitable for AV vendors to reinvent the autopilot design. It will make AVs capable of coping with the current dilemma that how the AVs make themselves more trustworthy in terms of safe travelling.

## Authors' contributions

FR and MN have equally contributed to the paper in both drafting the manuscript as well as revising it.

## Competing financial interests

The author(s) declare no competing financial interests.

## Data availability statement

All data generated or analyzed during this study are included in this published article.